\documentclass[10pt,twocolumn,letterpaper]{article}

\usepackage{wacv}              

\usepackage{graphicx}
\usepackage{amsmath}
\usepackage{amssymb}
\usepackage{booktabs}
\usepackage{algorithm}
\usepackage[noend]{algpseudocode}
\usepackage{setspace}       %
\usepackage{comment}
\usepackage{subcaption}
\usepackage{float}
\usepackage[accsupp]{axessibility}  

\newcommand{\Schurn}{S_\text{churn}}
\newcommand{\Stmin}{S_\text{tmin}}
\newcommand{\Stmax}{S_\text{tmax}}

\newcommand{\nsteps}{{n_\text{steps}}}
\newcommand{\rhomodel}{\rho_\text{model}}
\newcommand{\leftrhomodel}{\rho_\text{left,model}}
\newcommand{\rightrhomodel}{\rho_\text{right,model}}
\newcommand{\rhodataset}{\rho_\text{dataset}}
\newcommand{\leftrhodataset}{\rho_\text{left,dataset}}
\newcommand{\rightrhodataset}{\rho_\text{right,dataset}}
\newcommand{\smeq}{\mathord{=}}
\newcommand{\smgt}{\mathord{>}}
\newcommand{\xx}{\boldsymbol{x}}

\newcommand{\integrationSchema}{
\begin{figure*}
    \centering
    \includegraphics[width=1\linewidth]{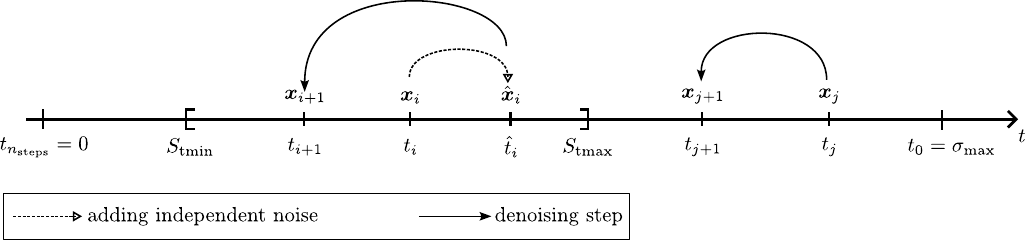}
    \caption{The integration scheme in \cite{edm}. In the window $[\Stmin, \Stmax]$ we first add independent noise to the image before performing denoising. Outside of the window (or when $~{\Schurn=0}$) we simply perform a deterministic denoising step. The denoising step is represented with a plain arrow, while the step of adding noise is represented with a dotted arrow. The denoising step can either be performed using Euler's method (calling once the network) or Heun's method (calling twice the network).}
    \label{fig:integration_schema}
\end{figure*}
}

\newcommand{\varGuidanceOne}{
\begin{figure}
    \centering
    \includegraphics[width=1\linewidth]{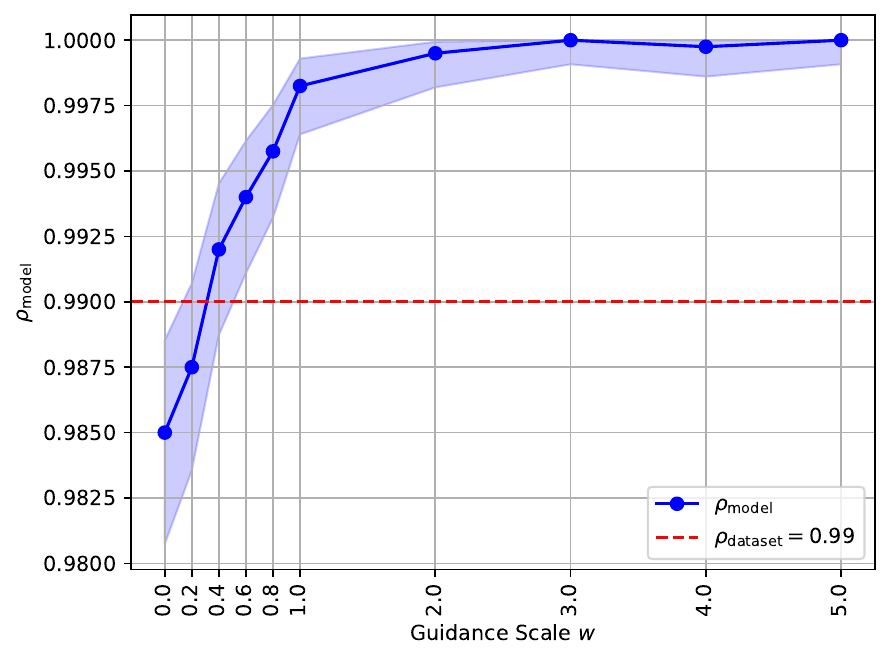}
    \caption{$\rhomodel$ vs guidance scale of CFG on 2-classes Biased MNIST ($\rhodataset\smeq0.99$) using Karras custom stochastic sampler.}
    \label{fig:25steps_var_guidance_until5}
\end{figure}
}

\newcommand{\SdVarNstepsRhoHpsEtaZero}{
\begin{figure}
    \centering
    \includegraphics[width=1\linewidth]{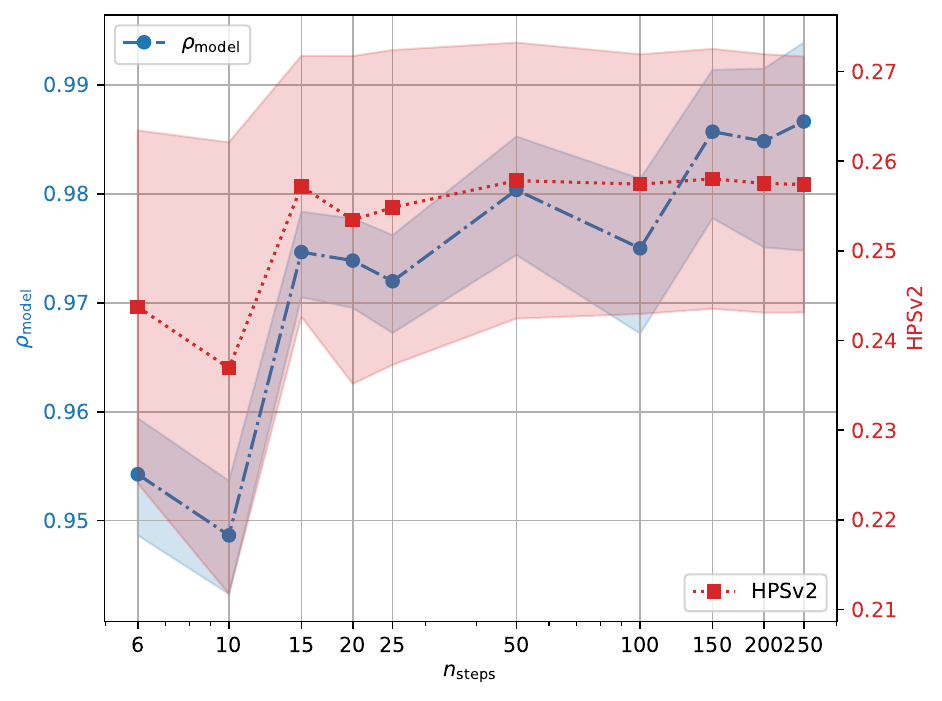}
    \caption{$\rhomodel$ and HPSv2 vs $\nsteps$ for Stable Diffusion v2.1 using DDIM sampler with $\eta\smeq 0$ (deterministic sampling).}
    \label{fig:SD_portraitlawyer_rho_hps_eta0_logscale}
\end{figure}
}

\newcommand{\TenClBmnistRhoVsSchurnVarNsteps}{
\begin{figure}
    \centering
    \includegraphics[width=1\linewidth]{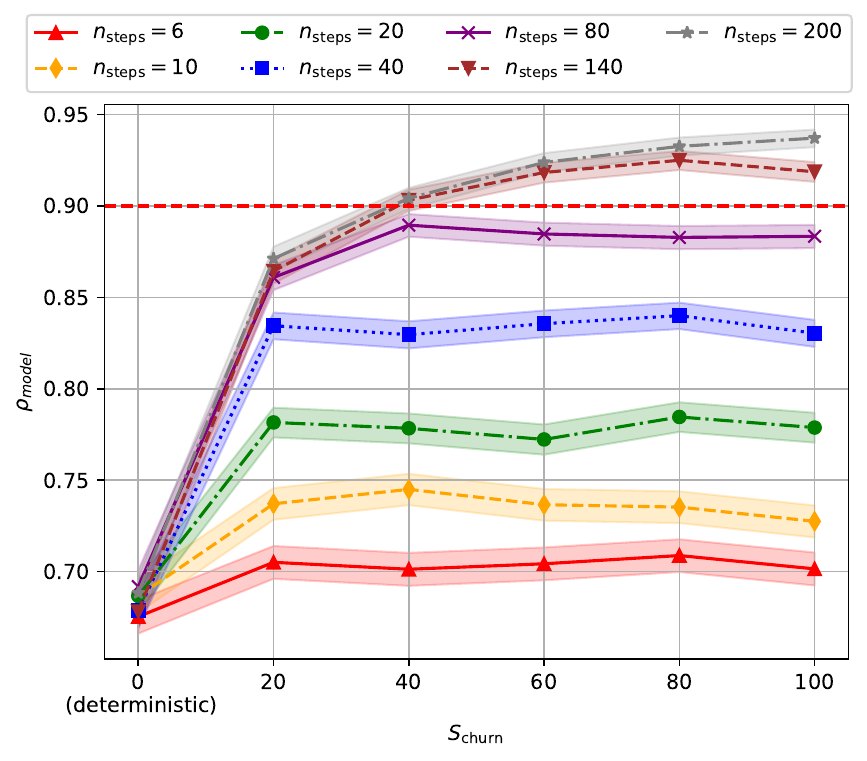}
    \caption{$\rhomodel$ vs $\Schurn$ for various $\nsteps$ for a model trained on 10-classes Biased MNIST ($\rhodataset\smeq 0.9$) using Karras stochastic sampler.}
    \label{fig:10clBmnist_rho0.9_vs_schurn_smin0.1_smax80_varnsteps}
\end{figure}
}

\newcommand{\varGuidanceBMNISTcdpm}{
\begin{figure}
    \centering
    \includegraphics[width=1\linewidth]{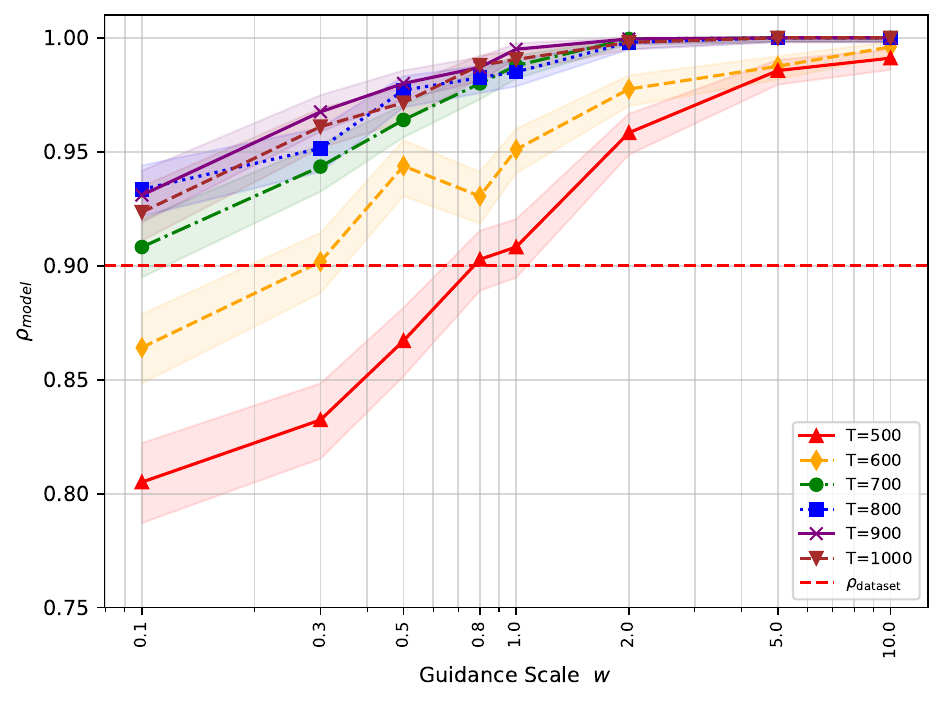}
    \caption{$\rhomodel$ vs guidance scale of CFG for various T on 2-classes Biased MNIST ($\rhodataset\smeq0.90$), standard DDPM sampler.}
    \label{fig:var_guidance_bmnist_cdpm}
\end{figure}
}

\newcommand{\test}{
\begin{figure*}
\centering
\begin{subfigure}{0.49\textwidth}
    \centering
    \includegraphics[width=1\linewidth]{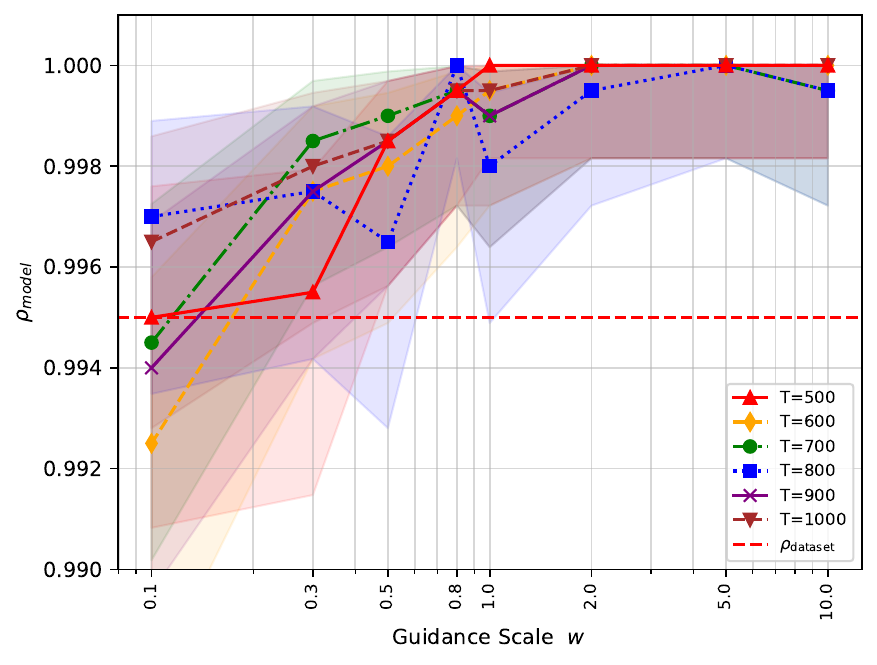}
    \caption{$\rhomodel$ vs guidance scale.}
    \label{fig:var_guidance_bffhq_cdpma}
\end{subfigure}
\begin{subfigure}{0.49\textwidth}
    \centering
    \includegraphics[width=1\linewidth]{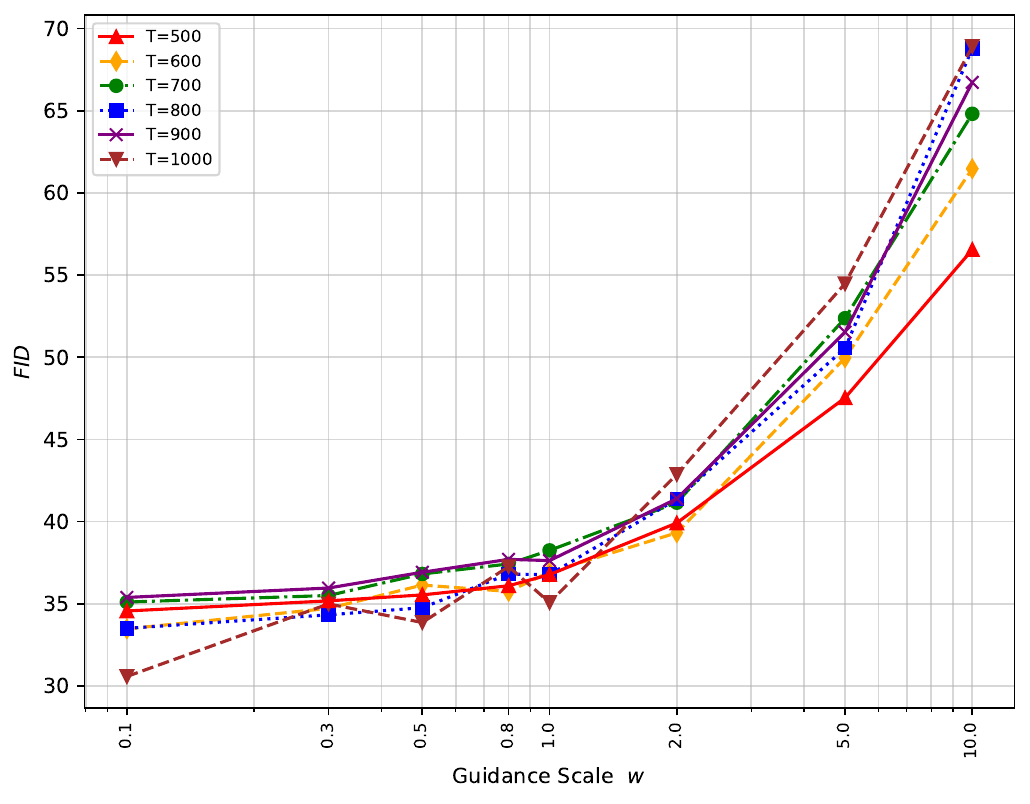}
    \caption{FID vs guidance scale.}
    \label{fig:fid_guidance_bffhq_cdpm}
\end{subfigure}
\caption{CFG on BFFHQ ($\rhodataset\smeq0.995$), standard DDPM sampler, varying the number of T.}
\end{figure*}
}

\definecolor{wacvblue}{rgb}{0.21,0.49,0.74}
\usepackage[pagebackref,breaklinks,colorlinks,allcolors=wacvblue]{hyperref}

\usepackage[capitalize]{cleveref}
\crefname{section}{Sec.}{Secs.}
\Crefname{section}{Section}{Sections}
\Crefname{table}{Table}{Tables}
\crefname{table}{Tab.}{Tabs.}

\def\Eq#1{Eq.~(\ref{eq:#1})}

\def\be{\begin{equation}}
\def\ee{\end{equation}}
\def\bea{\begin{eqnarray}}
\def\eea{\end{eqnarray}}

\def\sect#1{Section~\ref{sec:#1}}

\makeatletter
\DeclareRobustCommand\onedot{\futurelet\@let@token\@onedot}
\def\@onedot{\ifx\@let@token.\else.\null\fi\xspace}
\def\eg{{e.g}\onedot, }

\def\ie{{i.e}\onedot, }

\def\etal{\emph{et al}\onedot}

\definecolor{changes}{HTML}{0000Eb}

\def\wacvPaperID{1430} 
\def\confName{WACV}
\def\confYear{2026}

\begin{document}

\title{How I Met Your Bias: Investigating Bias Amplification in Diffusion Models}
\author{Nathan Roos$^{1}$ \quad Ekaterina Iakovleva$^{1}$ \quad Ani Gjergji$^{2}$ \quad Vito Paolo Pastore$^{2,3}$\thanks{Equal senior contribution.\\~\\This article has been accepted for publication at the IEEE/CVF Winter Conference on Applications of Computer Vision 2026 (WACV 2026).} \quad Enzo Tartaglione$^{1}$\footnotemark[1]\\
$^{1}$LTCI, Télécom Paris, Institut Polytechnique de Paris, France\\
$^{2}$MaLGa-DIBRIS, University of Genova, Italy\\
$^{3}$AIGO, Istituto Italiano di Tecnologia, Italy 
}
\maketitle


\begin{abstract}
Diffusion-based generative models demonstrate state-of-the-art performance across various image synthesis tasks, yet their tendency to replicate and amplify dataset biases remains poorly understood. Although previous research has viewed bias amplification as an inherent characteristic of diffusion models, this work provides the first analysis of how sampling algorithms and their hyperparameters influence bias amplification. We empirically demonstrate that samplers for diffusion models -- commonly optimized for sample quality and speed -- have a significant and measurable effect on bias amplification. Through controlled studies with models trained on Biased MNIST, Multi-Color MNIST and BFFHQ, and with Stable Diffusion, we show that sampling hyperparameters can induce both bias reduction and amplification, even when the trained model is fixed. Source code is available at \href{https://github.com/How-I-met-your-bias/how_i_met_your_bias}{https://github.com/How-I-met-your-bias/how\_i\_met\_your\_bias}.
\end{abstract}


\section{Introduction}
\label{sec:intro}

Diffusion-based probabilistic models have emerged as a leading paradigm in computer vision, consistently achieving state-of-the-art performance across a wide range of tasks. These tasks include, but are not limited to, unconditional image generation, class-conditional image synthesis, text-to-image synthesis, super-resolution, and image inpainting~\cite{cfg, edm, stable_diffusion_rombach, song_diffusion_sde}. The versatility and efficacy of these models underscore their transformative potential in both academic research and practical applications.

However, a known issue with these models is their tendency to replicate the biases in their training data. Indeed, these models are trained to fit the data distribution; therefore, if a bias is present in the data set, it is more than expected that the model will replicate it. However, what is at first glance astonishing is that in certain scenarios the bias has been observed to be amplified: this was mainly observed in demographic biases such as gender or skin color~\cite{quantifying_bias_tti_models, amplify_demographic_bias, survey_TTI_bias_evaluation, uncurated_image_text_datasets_shedding_light_on_demographic_bias}, but this issue potentially concerns any spurious relation present in the data~\cite{t2iat, DDB}.

\begin{figure}[t]
    \centering
    \includegraphics[trim={18 0 17 0},clip, width=0.95\columnwidth]{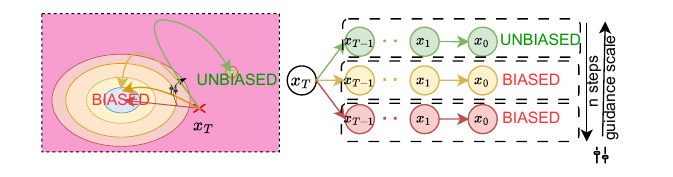} 
    \caption{In this work, we investigate the impact of the sampling parameters in (trained) diffusion models. \emph{Without retraining}, some parameters impact the generation of biased data.}
    \label{fig:teaser}
\end{figure}

However, previous studies tried to address this phenomenon without analyzing how the sampling process itself impacts the bias amplification phenomenon. Indeed, the most popular approaches to mitigate the impact of bias in the generation process, including attribute guidance~\cite{hong2023debiasing, parihar2024balancing} or a policy solver~\cite{he2024debiasing}, to name a few, all require retraining or finetuning the model.

In this work, our main goal is to empirically study the phenomenon of bias amplification in diffusion models from a sampling perspective. Unlike previous works that perceive bias as an intrinsic property of trained models, we view it as an adaptable behavior induced by the sampling procedure and reveal the factors that affect the generation of biased samples. We view our findings as complementary to the existing literature on bias mitigation, since adapting the sampler and its hyperparameters can be used in conjunction with other diffusion debiasing methods to further reduce the model bias in a training-free manner.

We summarize our contributions as follows:
\begin{itemize}
    \item We formalize the sampling process in diffusion models, identifying the core elements that can make the diffusion model deviate from a deterministic optimization trajectory (Sec.~\ref{sec:theoretical_diffusion_framewrok} and Sec.~\ref{sec:assessing_impact_parameters}).
    \item We hypothesize that introducing deviations from the deterministic trajectory would induce the generative model to generate more rare samples, allowing for a broader solution space exploration and de facto reducing the impact of bias (Sec.~\ref{sec:assessing_impact_parameters}). 
    \item We empirically validate our observations either in controlled scenarios or using state-of-the-art diffusion models with known biases (Sec.~\ref{sec:experiments}).
\end{itemize}

We believe that there are several practical applications of our findings. First, the proposed perspective calls for more careful construction of vanilla (\ie biased) baselines. Second, choosing specific samplers and tuning their hyperparameters can be used to reduce model bias without additional training. Finally, our findings propose an additional trade-off that can be taken into account when choosing hyperparameters to train a new diffusion model.

\section{Related works}
\label{sec:related works}

\textbf{Denoising Diffusion Probabilistic Model.} The \textit{denoising diffusion probabilistic model} (DDPM) was first introduced in Sohl-Dickstein~\etal~\cite{DDPM}. Here, the generation (later referred to as ``ancestral sampling") is performed by progressively denoising pure Gaussian noise into an image by following a Markov Chain, with a Gaussian transition kernel parameterized by a neural network. Its main drawback is that the forward pass through the network is required at each denoising step, resulting in thousands of evaluations required to generate a single batch of images. Song~\etal~\cite{song_diffusion_sde} later linked DDPMs to stochastic differential equations (SDEs) and ordinary differential equations (ODEs) by designing DDPMs with infinitesimal time steps. In this work, the authors demonstrate that SDEs can be used to formulate both score-based models and DDPMs with varying discretizations. Finally, Karras~\etal~\cite{edm} proposed a reformulation of the continuous framework to improve the interpretability of hyperparameter selection and to unify existing methods. 
In~\cite{dhariwal2021diffusion}, the quality of images generated with diffusion models is improved by conditioning the diffusion process on the class label using an external classifier guidance. Specifically, by incorporating gradients from an external classifier during sampling, the authors achieved a new state-of-the-art in image synthesis, outperforming GANs. Later, in~\cite{ho2022classifier}, the authors propose to avoid the use of an external classifier, showing how improved image synthesis can be obtained in a classifier-free setting, where an unconditional and a conditional diffusion model are jointly trained, with their score combined to achieve performance similar to the classifier-guidance model. \\
\textbf{Improving sampling.} 
On the one hand, diffusion models achieve better image synthesis than GANs. However, as the generative process may involve thousands of steps (being the reverse of the forward diffusion process), the computational cost in terms of time and resources is significantly higher than that of GANs. Recently, several works have investigated the problem of high training cost and long inference time for diffusion models.
In this context, Rombach~\etal~\cite{stable_diffusion_rombach} propose to use \textit{latent diffusion models} (LDMs), which are applied in the latent space of pre-trained autoencoders, thus considerably reducing the training and inference cost. Stable Diffusion~\cite{stable_diffusion_rombach} belongs to this class of models.
Importantly, a large body of literature has been dedicated to studying sampling in diffusion models with the goal of decreasing its computational cost while preserving or increasing the quality of the generated samples and enabling control over the generated output. \\
To decrease the number of generation steps, Song~\etal~\cite{DDIM} propose the DDIM (denoising diffusion implicit model) sampler which improves the DDPM sampler \cite{DDPM} while using only certain steps of the denoising Markov chain. It facilitates going from $1000$ required steps to as few as $20$ without significantly compromising quality. \\
Methods that work with \textit{exposure bias}~\cite{ning2023input,ning2023elucidating,yu2024unmasking,zhou2024dream,zhanganti,yu2025frequency} in diffusion models aim to reduce the discrepancy between the training regime and the inference. In the former, the noise predictor relies on the ground truth samples, while in the latter, it relies on the self-generated samples. This results in cumulative error and sampling drift as the number of inference steps increases. Li and Qu~\cite{timeshiftsampler} propose to mitigate this exposure bias by introducing the Time-Shifted Sampler, which dynamically aligns intermediate samples with the noise levels they best match. 
These works aim to improve the \textit{quality} of the generated images rather than the \textit{distributional coverage} of the image space. In this paper, we are interested in analyzing the impact of the sampling procedure on the generated images from a bias perspective.\\
\textbf{Bias amplification in diffusion models.} The tendency of diffusion models to replicate or amplify biases present in the training data is a well-known phenomenon in the literature. Many works~\cite{quantifying_bias_tti_models, auditing_internal_bias_dynamics, t2iat, amplify_demographic_bias, survey_TTI_bias_evaluation, uncurated_image_text_datasets_shedding_light_on_demographic_bias} observe it both in large text-to-image models (e.g., Stable Diffusion~\cite{stable_diffusion_rombach} or DALL-E~\cite{ramesh2021zero})  and in smaller specialized models~\cite{DDB}. Some works even rely on the hypothesis that diffusion models amplify biases to build debiasing pipelines for classifiers~\cite{DDB}. However, we argue that the term ``bias amplification'' should be interpreted cautiously, since it assumes that there is a baseline for comparison, and this baseline is not always grounded in the training data. For example, Bianchi~\etal~\cite{amplify_demographic_bias} highlight social stereotypes in the outputs of Stable Diffusion by comparing them to real-world statistics, but this does not indicate whether the model has amplified or merely replicated the biases in its training data. This distinction is critical for understanding how diffusion models work, as well as for determining whether bias mitigation should focus on model training and sampling, or curation of the training data.\\
Some works observe bias amplification with respect to the data set. Ciranni~\etal~\cite{DDB} reported that a diffusion model trained on the Waterbirds dataset amplified the bias, using human annotators to quantify the bias in the generated images. Seshadri~\etal~\cite{bias_amplification_paradox} compare the gender ratio in images generated by Stable Diffusion with the gender ratio of the data set. They highlight that even though it might seem that Stable Diffusion amplifies gender stereotypes, much of it can be explained by the distributional shift between the detailed captions that the model was trained on and the concise gender-neutral prompts used in the evaluation. Accounting for this shift eliminates much of the observed bias amplification, though some of it remains.\\
More recent work on bias mitigation in diffusion models approaches the problem from the perspective of mechanistic interpretability~\cite{shi2025dissecting}, the properties of text embedding space~\cite{kuchlous2024bias,kim2024discovering,li2025responsible}, and model architecture~\cite{hakemi2025deeper}. All these works treat the pair of the noise prediction network and the sampler jointly and statically, assuming the strength of the bias to be a constant property. In our work, we attempt to decouple the denoiser and sampler and study the role of the latter in bias manipulation.\\
To the best of our knowledge, none of the existing studies on bias have investigated whether the observed bias amplification depends on the sampling procedure and its hyperparameters. In this regard, our work is novel and thus has no baseline with which to compare.

\section{Problem statement}
\label{sec:problem_statement}

In this section, we first formalize our definition of bias in Sec.~\ref{sec:bias_formalization}. Then, in Sec.~\ref{sec:theoretical_diffusion_framewrok}, we give an overview of the theoretical aspects of diffusion models that impact the inference procedure. In Sec.~\ref{sec:assessing_impact_parameters}, we discuss the role of sampling techniques and their hyperparameters on bias amplification. Finally, in Sec.~\ref{sec:naive_mc_estimator} we describe the bias estimator that we use in our work.

\subsection{Definition of bias}
\label{sec:bias_formalization}

Let $X$ be a random variable sampled from the image distribution $\mathcal{X}$, and let $Y \in \mathcal{Y}$ be the corresponding target attribute given to the generative model as a condition. Finally, let $B \in \mathcal{B}$ be an attribute spuriously correlated with $Y$ in the given data distribution $\mathcal{D} = (\mathcal{X}, \mathcal{Y})$. We then call $B$ the \textit{bias} attribute. For simplicity, we assume that $\mathcal{Y}$ and $\mathcal{B}$ are finite and that each target attribute $Y$ is associated with a unique bias attribute $B$. For example, $X$ could be the image of a person, $Y$ might denote the profession of the depicted person, and $B$ could represent their gender.\footnote{We recognize that gender is complex and multifaceted, and although we do not seek to reinforce a binary view of gender, this research considers men and women in a binary sense to simplify the analysis.} Then in the data, the value $~{y_1=\text{``\texttt{surgeon}"}}$ might be correlated with the value $~{b_1=\text{``\texttt{male}"}}$, and the value $~{y_2=\text{``\texttt{nurse}"}}$ might be correlated with the value $~{b_2=\text{``\texttt{female}"}}$. 

We adopt the vocabulary of the debiasing literature~\cite{DDB, DFA} and call \textit{bias-aligned} images whose target attribute aligns with their bias attribute (\eg male surgeons and female nurses) and call \textit{bias-conflicting} images whose target attribute does not align with their bias attribute (\eg female surgeons and male nurses). Additionally, the ``amount" of bias in a given distribution (\ie, the difficulty of the debiasing task) is usually quantified by the parameter $\rho$, which is defined as the probability that the sample is bias-aligned~\cite{DDB, ReBias}.
In a given data set, the parameter $\rhodataset$ is simply the ratio of bias-aligned samples. 
In \sect{naive_mc_estimator} we discuss how we estimate the bias $\rhomodel$ in the output distribution of the generative model. 

Although other metrics to quantify bias in generative models have been proposed in the literature~\cite{quantifying_bias_tti_models, auditing_internal_bias_dynamics, t2iat}, they have not been widely adopted because they are less interpretable or harder to compute. 
This motivates our choice of $\rho$ as a metric that is easy to compute and interpret and is well-known in the literature~\cite{bias_amplification_paradox, amplify_demographic_bias, t2iat}.

\subsection{Preliminaries of diffusion models}
\label{sec:theoretical_diffusion_framewrok}

\integrationSchema

\textbf{Discrete probabilistic models.} 
We consider DDPM~\cite{DDPM}, a diffusion model that learns to reverse a noising process described by a discrete Markov chain with $T$ steps. It progressively corrupts the original data sample $x_0$ with gradually increasing Gaussian noise according to some pre-defined variance schedule $0 < \sigma_1, \dots, \sigma_{T} < 1$. For step $t \in [1, T]$:
\begin{equation}
    x_{t} = \sqrt{\alpha_{t}} x_{t-1} + \sigma_{t}\epsilon,
\end{equation}
Where $\alpha_{t} = 1 - \sigma_t^2$ and $\epsilon \sim \mathcal{N}(0, I)$. Ho~\etal propose reparameterization $x_{t}(x_{0}, \epsilon) = \sqrt{\overline{\alpha_t}}x_{0} + \sqrt{1 - \overline{\alpha_t}}\epsilon$ with $\overline{\alpha}_{t} = \prod_{s=1}^{t} \alpha_s $. In the reverse variational Markov process, a neural network $\epsilon_{\theta}(x, t)$ parameterized with $\theta$ is trained to approximate the true noise $\epsilon$. Samples of the learned data distribution are obtained iteratively, where each step $t = T, \ldots, 1$ looks as follows:
\begin{equation}
    x_{t-1} = \frac{1}{\sqrt{\alpha_{t}}}\left(x_{t} - \frac{1-\alpha_{t}}{\sqrt{1-\overline{\alpha_{t}}}}\epsilon_{\theta}(x, t)\right) + \sigma_{t} \epsilon.
\end{equation}

\noindent\textbf{Continuous diffusion models.} We consider EDM~\cite{edm}, a continuous diffusion framework with a stochastic sampler that adopts the probability-flow ODE formulation of diffusion models from~\cite{song_diffusion_sde} and sets $\alpha_{t}=1$ and $\sigma(t)=t$:
\begin{equation}
    d\xx = - \dot{\sigma}(t)\sigma(t)\nabla_{\xx} \log p(\xx;\sigma(t))dt.  \label{eq:ode}
\end{equation}
Here $~{\nabla_{\xx} \log p(\xx;\sigma(t))}$ is the \textit{score} function denoted as $s_\theta(\xx;\sigma)$. Karras~\etal~\cite{edm} show that it can be expressed in the following way: 
\begin{equation}
    \nabla_{\xx} \log p(\xx;\sigma) = \frac{D(\xx;\sigma) - \xx}{\sigma^2}, \label{eq:score_denoising}
\end{equation}
where $\xx$ is a sample of the distribution $~{p(\xx;\sigma)}$, and $~{D(\xx;\sigma)}$ is the denoising function. A neural network $D_\theta$ parameterized with $\theta$ is trained to approximate the true $D$, which is further used in \cref{eq:score_denoising} to estimate the score function $~{s_\theta(\xx;\sigma)}$ in \cref{eq:ode}. 

In EDM~\cite{edm}, samples of the learned data distribution $p_\theta$ are obtained by integrating \Eq{ode} from $\sigma_\text{max}$ to $0$ in $\nsteps$ integration steps using a custom stochastic sampler. 
At each time step $t$, if $t \in [\Stmin,\Stmax]$, an independent noise, referred to as \textit{fresh noise}, is added to the current image, with the variance controlled by the hyperparameter $\Schurn$. $\Schurn~{=0}$ corresponds to deterministic sampling. After that, a denoising step is made from this image using an ODE solver. The process is illustrated in Fig.~\ref{fig:integration_schema}.

\noindent\textbf{Conditional sampling.} The simplest way to generate an image from an arbitrary class with a single diffusion model is to use classifier-free guidance (CFG) \cite{cfg}. CFG introduces a flexible conditioning approach that balances conditional and unconditional outputs without relying on an external classifier. During training, the model is randomly conditioned on a special unconditional class identifier $\varnothing$ with some predefined probability $p_\text{uncond}$, which is a hyperparameter. This enables the model to learn both generation modalities. During sampling, a guidance scale $w$ modulates the influence of conditioning, with the guided score prediction being defined as follows:
\begin{align}
    \tilde{s}_\theta(\xx;\sigma;y) = (1+w)s_\theta(\xx;\sigma;y) - w \cdot s_\theta(\xx;\sigma;\varnothing)& \notag \\
    =s_\theta(\xx;\sigma;y) + w\cdot (s_\theta(\xx;\sigma;y) - s_\theta(\xx;\sigma;\varnothing)),& \label{eq:CFG}
\end{align}
where $\xx \in \mathcal{X}$ is an image, $y\in\mathcal{Y}$ is a class label, $s_\theta(\xx;\sigma;y)$ is the score predicted by the model parameterized with $\theta$ at the noise level $\sigma$ and conditioned on the class $y$, and $s_\theta(x;\sigma;\varnothing)$ is the corresponding unconditional score.
The unconditional score prediction can be obtained either by conditioning the network with the null token $\varnothing$ or by aggregating all conditional scores: 
\begin{equation}
    s_\theta(\xx;\sigma;\varnothing) = \sum_{y \in \mathcal{Y}} s_\theta(\xx;\sigma;y) \cdot p(y), \label{eq:aggregated_score_pred}
\end{equation}
where $p(y)$ is the probability of class $y$ in the training set.

\subsection{Sampling hyperparameters and bias-correlation in generated images}
\label{sec:assessing_impact_parameters}

Recent works~\cite{edm, DDIM, timeshiftsampler} have proposed sampling strategies alternative to the original formulation in~\cite{DDPM}, to reduce the time needed for generation. 
The commonly used sampling strategies are characterized by a set of hyperparameters, which can significantly influence the generation process~\cite{ho2022classifier}. However, to our knowledge, no work has critically analyzed the impact of such parameters on the bias-correlation for diffusion-generated images when the training set is strongly biased.  
To fill this gap, we investigate the extent to which bias is amplified (or reduced) in diffusion models, depending on the choice of sampling hyperparameters. 
Specifically, our objective is to determine whether bias amplification, which has been observed in recent works~\cite{d2024openbias, DDB}, depends on the specific set of sampling parameters used in generation. 
To structure our analysis, we categorize the sampling hyperparameters into three groups:
\begin{itemize}
    \item conditioning strength: $\{w\}$,
    \item computational cost: $\{\nsteps, \text{type of integration scheme}\}$,
    \item amount of stochasticity: $\{\Schurn \text{ or }\eta, \Stmin, \Stmax\}$.
\end{itemize}
\textbf{DDPM.} In the original implementation of DDPM~\cite{DDPM}, the generation framework is discrete. As such, the only parameter affecting generation quality (and bias, in our analysis) is the number of diffusion steps. \\
\textbf{CDPM.} In Conditional DPMs, the sampling parameters include the number of diffusion steps and the \textit{conditioning strength} (see Sec.~\ref{sec:theoretical_diffusion_framewrok}). The latter governs the extent to which the model adheres to the provided prompt or class label during sampling. In our experiments, this is controlled by the classifier-free guidance (CFG) scale parameter, denoted by $w$. Increasing the value of $w$ typically results in a stronger alignment with the conditioning signal, with improved quality and decreased diversity~\cite{ho2022classifier}. We hypothesize that higher guidance scales may lead to greater bias amplification, as CFG has been observed to steer generated samples toward average representation of the target class (often corresponding to bias-aligned samples).\\
\textbf{Continuous sampling framework.} Here, the sampling parameters include the conditioning strength and the ones associated with the \textit{computational cost}. The latter primarily influences the number of calls to the network (NFE), being the most computationally expensive component of the sampling process. In particular, two factors are relevant: the integration scheme and the number of integration steps, denoted by $\nsteps$. 
\\The choice of integration scheme affects the number of network evaluations required per step (for example, Euler's method requires one evaluation per step, while Heun's method requires two). These hyperparameters are of interest not only for their impact on efficiency but also because they influence the numerical error introduced during sampling. Importantly, integration errors can introduce additional variability in the generation process, which in turn could lead to bias reduction.
Finally, the hyperparameters related to the \textit{amount of stochasticity} determine when and to what extent the sampling trajectory can deviate from a deterministic path. We consider two representative cases. 
In this work, we evaluate the popular continuous sampling framework proposed by Karras~\etal~\cite{edm} with three different samplers: EDM-sampler \cite{edm}, VP-sampler \cite{song_diffusion_sde}, and DPM-Solver \cite{dpm_solver}. The results of the EDM-sampler are discussed in the main paper, while the other two are left to Appendix C.
Here, ${\Stmin, \Stmax}$ specifies the time window within which additional noise is injected, while $\Schurn$ controls the variance of the added noise. For other hyperparameters, we use the values from \cite{edm}. Second, in the case of the DDIM sampler applied to Stable Diffusion, the parameter $\eta$ modulates the variance of the noise introduced during sampling. 
In this context, the effect of increased stochasticity on bias-correlation for generated images is largely unexplored.

\subsection{Na\"ive Monte-Carlo estimator of model bias}
\label{sec:naive_mc_estimator}

Studying the impact of sampling parameters on bias-target correlation in generated images requires a rigorous definition of a bias-measuring protocol. 
In this section, we present the mathematical formulation for estimating the parameter $\rho$ introduced in \sect{bias_formalization}. We use this estimator in our experiments as an evaluation metric.  

The na\"ive Monte-Carlo estimator $\hat{\rho}_{model}$, commonly used in the literature, consists of generating many samples from the model conditioning on a given class and then determining the ratio of images correlated with the bias attribute (the \textit{bias-aligned} samples) using an oracle (\eg a human annotator, a CNN classifier, a captioner or Visual Question Answering (VQA) model~\cite{bias_amplification_paradox, auditing_internal_bias_dynamics, DDB, survey_TTI_bias_evaluation}). 

Let us assume that our objective is to measure the level of bias for a single target attribute $~{y \in \mathcal{Y}}$ and its correlated bias attribute $~{b \in \mathcal{B}}$. We denote by $~{f_b : \mathcal{X} \longrightarrow\{0,1\}}$ the oracle that determines whether a sample $\xx$ is bias-aligned ($f_b(\xx)~{=1}$) or bias-conflicting ($f_b(\xx)~{=0}$).

The estimator is computed as follows: sample independently $N$ images $~{\xx_1^y, \cdots, \xx_N^y \sim p_\theta(\xx|y)}$ from the model $\theta$ conditioning on $y$ and compute:
\begin{equation}
    \hat{\rho}_{model}^y = \frac{1}{N} \sum_{i=1}^N f_b(\xx_i^y).
\end{equation}

To compute the estimator for $K$ target attributes $~{(y_1, \cdots,y_K)}$ with associated bias attributes $~{(b_1, \cdots,b_k)}$, we independently sample $N$ images from each of the $K$ classes: $~{\xx_1^{y_1}, \cdots, \xx_N^{y_1} \sim p_\theta(\xx|y_1)}$, $\cdots$, $~{\xx_1^{y_K}, \cdots, \xx_N^{y_K} \sim p_\theta(\xx|{y_K})}$ and compute:
\begin{equation}
    \hat{\rho}_{\text{model}} = \frac{1}{KN} \sum_{k=1}^K \sum_{i=1}^N f_{b_k}(\xx_i^{y_k}). \label{eq:naive_multitarget_rho_est}
\end{equation}

We can interpret $\rhomodel$ as the parameter of a binomial distribution (where sampling a bias-aligned sample is a success). Thus, we can obtain confidence intervals on $\hat{\rho}_{model}^y$  and $\hat{\rho}_{model}$ by using the conservative Clopper-Pearson method \cite{clopper_pearson}.

\section{Experiments}
\label{sec:experiments}

After presenting at a glance the employed setup (Sec.~\ref{sec:experimental_setup}), Sec.~\ref{sec:results} presents the main results for all of the sampling parameters identified in Sec.~\ref{sec:assessing_impact_parameters}. A discussion of the observed phenomena is provided in Sect.~\ref{sec:discussion}. 

\subsection{Setup}
\label{sec:experimental_setup}

\textbf{Datasets.} For our experiments, four datasets have been employed. \textit{Biased MNIST} \cite{ReBias} is a variant of the MNIST dataset used to evaluate debiasing methods. It introduces bias by associating specific background colors with digit classes, with a proportion of samples having matching background colors and the rest having randomly assigned colors.  \textit{Multi-Color MNIST} \cite{multicolor_mnist} is another variant of MNIST. It introduces two biases by colouring the left and right halves of the background. For the sake of conciseness, its results are left to Appendix C. \textit{BFFHQ} \cite{biaswap_bffhq} is a dataset of face images that introduces demographic biases, correlating age labels with gender. A significant majority (95\%) of images labeled as ``young'' depict female subjects, while the same proportion of "old" images depict male subjects. \textit{LAION-5B}~\cite{laion_5b} is a large-scale dataset of image-caption pairs collected from the web, reflecting societal stereotypes. The focus is on gender bias in the context of the occupation ``lawyer''. Estimating the baseline value of bias in LAION-5B is complex due to distributional shifts between training captions and inference prompts. Thus, the study focuses on changes in bias with different sampling hyperparameters. A more detailed description of the datasets can be found in Appendix A.

\noindent \textbf{Architectures.} On Biased MNIST and BFFHQ, we train diffusion models with the UNet~\cite{UNet} architecture. We also use the preconditioning proposed by Karras~\etal~\cite{edm}. Images are resized to 32$\times$32 for Biased MNIST and 64$\times$64 for BFFHQ. We trained our model on BFFHQ on 60 million images on a single NVIDIA A40 GPU. Stable Diffusion is a text-to-image model trained on LAION-5B that generates images based on a textual description (prompt)~\cite{stable_diffusion_rombach}. The version we use is Stable Diffusion v2.1-base, generating images of resolution 512$\times$512. \\
\textbf{Bias oracle.} To determine the bias attribute of the images generated by models trained on Biased MNIST, we compute the closest color to the average color of the non-white pixels. Regarding Stable Diffusion and the model trained on BFFHQ, we need to infer the gender of the individual in the images. To this end, we prompt the multimodal LLaVA-v1.5-7B model~\cite{LLaVA1.5} with the images and the text prompt ``\textit{Is the person a male? (answer only with yes or no)}". The model is allowed to generate a single new token. The images are kept only if the answer is ``yes" or ``no".\\
Additional details, including information about the network architecture, image generation, image selection, and image quality assessment, can be found in Appendix B.

\test

\varGuidanceOne

\varGuidanceBMNISTcdpm

\subsection{Results}
\label{sec:results}
\textbf{Effect of diffusion timesteps.}
First of all, we analyze the impact of diffusion timesteps (T), notoriously influencing the generation process in DDPMs~\cite{DDPM}.
In standard DDPM, changing T does not amplify the bias present in BFFHQ (\cref{fig:var_guidance_bffhq_cdpma}).
Increasing the number of T has been shown to result in more precise reverse diffusion, yielding higher quality samples (\cref{fig:fid_guidance_bffhq_cdpm}, where
$~{T=1000}$ is the typically employed value).\\
\textbf{Effect of conditioning strength.}
As we study the effect of the sampling hyperparameters on the amplification of bias, the one we should study first is the guidance scale $w$ in CFG~\cite{cfg}. As anticipated in~\cref{fig:25steps_var_guidance_until5,fig:var_guidance_bffhq_cdpma,fig:var_guidance_bmnist_cdpm}, we observe a positive correlation between the guidance scale $w$ and the level of bias of the model $\rhomodel$. Notably, we identify two distinct regimes: at lower values of $w$ and fewer timesteps, $\rhomodel$ falls below the dataset level of bias $\rhodataset$, indicating a \textit{reduction} of bias during sampling. In contrast, at higher values of $w$, the model exhibits bias \textit{amplification} with $~{\rhomodel\smgt\rhodataset}$. 

\begin{figure*}
\centering
\begin{subfigure}{0.45\textwidth}
    \centering
    \includegraphics[width=1\linewidth]{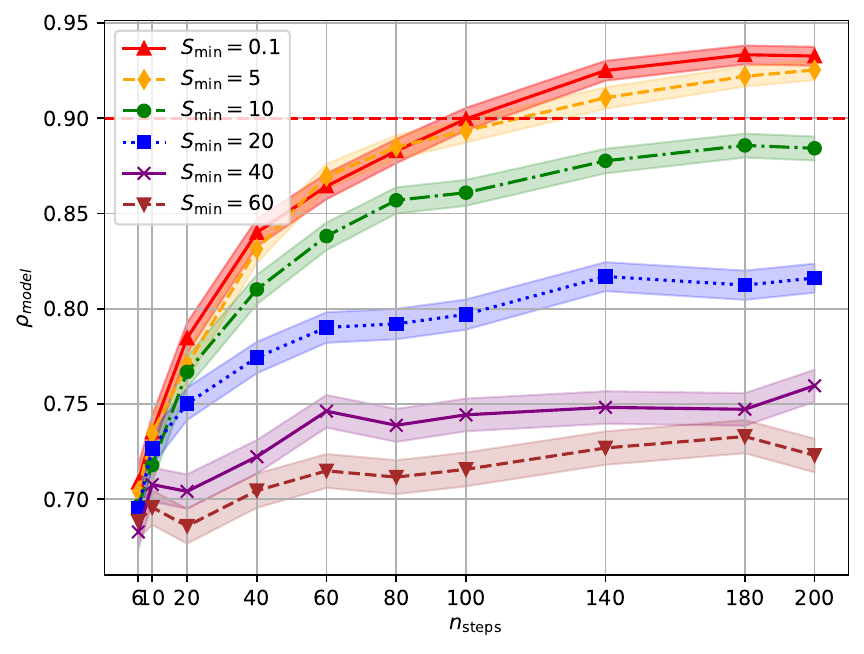}
    \caption{Study on $\Stmin$.}
    \label{fig:10clBmnist_rho0.9_vs_nsteps_schurn80_smax80_varsmin}
\end{subfigure}
~
\begin{subfigure}{0.45\textwidth}
    \centering
    \includegraphics[width=1\linewidth]{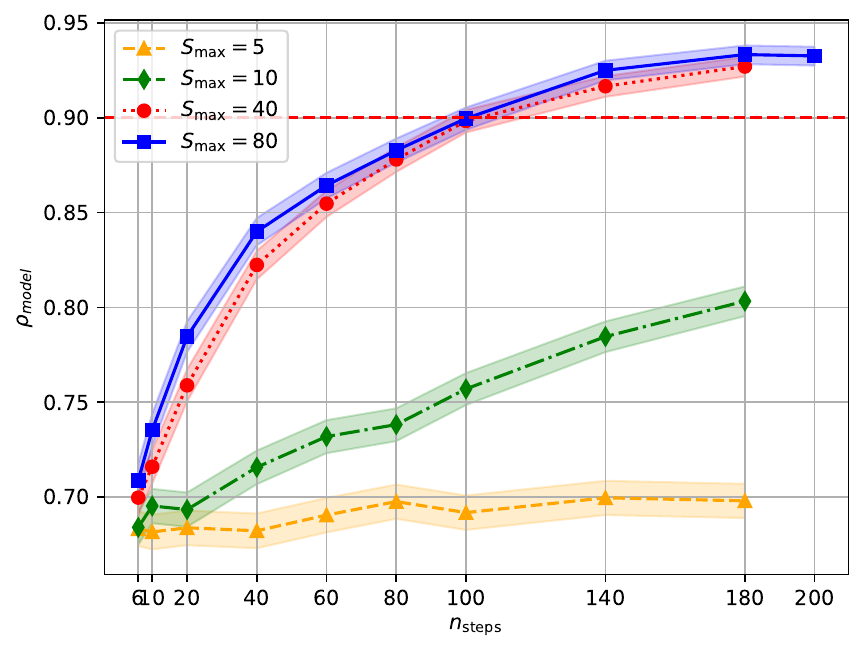}
    \caption{Study on $\Stmax$.}
    \label{fig:10clBmnist_rho0.9_vs_nsteps_schurn80_smin0.1_varsmax}
\end{subfigure}
\caption{$\rhomodel$ vs $\nsteps$ for a model trained on 10-classes Biased MNIST ($\rhodataset\smeq 0.9$) using Karras stochastic sampler.}
\end{figure*}

\noindent\textbf{Effect of the number of integration steps.}
We vary the number of sampling steps $\nsteps$ for models trained on Biased MNIST and BFFHQ and for Stable Diffusion.
In 10-classes Biased MNIST, $\rhomodel$ increases as $\nsteps$ increases, showing bias \textit{reduction} at low $\nsteps$ followed by bias \textit{amplification} at high $\nsteps$. The same pattern appears in BFFHQ and Stable Diffusion (\cref{fig:SD_portraitlawyer_rho_hps_eta0_logscale}).
For Stable Diffusion, $\rhomodel$ varies significantly with $\nsteps$ (from $0.949$ to $0.987$)—a previously unexplored phenomenon. Notably, the HPSv2 score initially increases with $\nsteps$ but plateaus quickly, while $\rhomodel$ continues increasing.

\SdVarNstepsRhoHpsEtaZero

\noindent\textbf{Time window with fresh noise.} We first observe that the fresh noise does not have the same effect depending on the time window in which we inject it. In \cref{fig:10clBmnist_rho0.9_vs_nsteps_schurn80_smin0.1_varsmax}, we observe almost no difference between the results obtained with $~{[\Stmin\smeq0.1,\Stmax\smeq40]}$ and $[0.1,80]$, which suggests that the noise injected during the time window $[40, 80]$ plays no role in the amplification of the bias. This is confirmed by \cref{fig:10clBmnist_rho0.9_vs_nsteps_schurn80_smax80_varsmin} where we see that the results obtained with $[40,80]$ and $[60,80]$ are close to each other. Similarly, in \cref{fig:10clBmnist_rho0.9_vs_nsteps_schurn80_smax80_varsmin} the results between $[0.1,80]$ and $[5,80]$ are close, which suggests that the noise injected during the time window $[0,5]$ does not affect the amplification of bias. 

\TenClBmnistRhoVsSchurnVarNsteps

\noindent\textbf{Variance of the fresh noise.} We experiment with the variance of the injected fresh noise by varying $\Schurn$. In \cref{fig:10clBmnist_rho0.9_vs_schurn_smin0.1_smax80_varnsteps}, we see that only once $\Schurn>0$ (\ie the sampling is stochastic rather than deterministic), there is an effect of $\nsteps$. Results in Appendix C further validate that $\rhomodel$ is positively correlated with $\Schurn$.

\subsection{Discussion}
\label{sec:discussion}

\textbf{Numerical errors help in debiasing.} Our experiments highlight that all three types of identified hyperparameters impact the phenomenon of bias amplification. 
The \textit{conditioning strength} is expected to affect the generated distribution and bias amplification. More specifically, high guidance scales cause samples to resemble class averages (bias-aligned samples), reflecting the known quality-diversity tradeoff in CFG.
The \textit{number of sampling steps} can also dramatically reduce or amplify bias. We hypothesize that a few integration steps in the sampler cause numerical errors in early timesteps, steering samples away from the learned distribution. Later time steps correct these errors, bringing samples toward the learned data distribution while exploring the rare solutions.
The \textit{stochasticity} in the denoising impacts bias amplification only during the specific time window when bias features are being decided.
Given the prevalence of bias-aligned samples in both data and generation, most of the sampling trajectories from the prior distribution $~{\mathcal{N}(0,\sigma_\text{max})}$ to the data distribution are likely "bias-leading trajectories".
When noise is injected, there is a high probability that we will jump between the trajectories and land on those that are biased. 
Assuming jumps from non-bias-leading to bias-leading trajectories are more probable than the reverse ones, this hypothesis can explain observations made in this work. 
Specifically, the more sampling steps there are with stochasticity enabled, the more likely it is to be on a bias-leading trajectory.

\noindent\textbf{What distribution did the model learn?} A crucial takeaway from our experiments is that \textit{the generated distribution is clearly dependent on the sampling hyperparameters}. It raises the following fundamental question: which choice of sampling hyperparameters reflects the true distribution that the model has learned? We see that, depending on the choices of sampling steps, we observe either amplification or reduction of bias \textit{showcased by the same model}. We could argue that the distribution learned by the model is the one obtained when exactly integrating the denoising equations \cref{eq:ode}. This means that the true distribution learned by the model is close to that obtained with a very large $\nsteps$. 

\noindent\textbf{Towards debiasing through sampling.} Looking at our results, it appears evident that the reduction of bias that we observed is an opportunity to devise a debiasing sampling strategy. Following this, one idea is to exploit the decoupling between $\rhomodel$ and the quality metric HPSv2 observed in \cref{fig:SD_portraitlawyer_rho_hps_eta0_logscale}. Indeed, at some point, the quality stops augmenting, while $\rhomodel$ does. This shows that it might be possible to design a proper debiasing strategy that would not require retraining of the diffusion model. 

\noindent\textbf{Limitation.}
The major limitation of this work lies in the fact that there are potentially a large number of possible factors or combinations of factors that contribute to bias amplification. In some experiments, we had to fix all hyperparameters except for one under investigation, potentially missing higher-order interactions between the hyperparameters. 

\section{Conclusion}
\label{sec:conclusion}

In this work, we grounded and framed the problem of biases in diffusion models and we empirically showed that the choice of sampling hyperparameters plays a crucial role in the evaluation of biases in diffusion models: some lead to \emph{amplification} of bias, while others -- to bias \emph{reduction}. This observation is confirmed at three different scales of data set complexity and in different diffusion models. 

Our results pave the way for two research aspects to be explored in future work. First, it is in principle possible to design a debiasing strategy for diffusion models based on made observations. Simply changing the sampling parameters reduces the impact of the bias, however, it is unable to fully eradicate it. Second, a deeper theoretical interpretation of the observed phenomenon could, in principle, facilitate a better understanding of the underlying mechanisms in optimization of diffusion models. 

\\\\
\noindent \textbf{Acknowledgments.} We extend our sincere gratitude to Rémi Nahon and Stephan Alaniz for their precious assistance during the research that led to this work. This work was supported in part by the French National Research Agency (ANR) in the framework of the JCJC project “BANERA” under Grant ANR-24-CE23-4369, and in part by the Hi!PARIS Center on Data Analytics and Artificial Intelligence. 

\newpage

\appendix

\section{Datasets}

\begin{table}[b]
  \centering
  \resizebox{\columnwidth}{!}{
  \begin{tabular}{@{}ccc@{}}
    \toprule
    Digit class & RGB values of associated color & Color\\
    \midrule
    0 & $(255,0,0)$ & red \\
    1 & $(0,255,0)$ & green \\
    2 & $(0,0,255)$ & blue \\
    3 & $(255,255,0)$ & yellow \\
    4 & $(255,0,255)$ & magenta \\
    5 & $(0,255,255)$ & cyan\\
    6 & $(255,128,0)$ & orange\\
    7 & $(255,0,128)$ & rose\\
    8 & $(128,0,255)$ & electric violet\\
    9 & $(128,128,128)$ & grey\\
    \bottomrule
  \end{tabular}
  }
  \caption{RGB values of the colors associated with the digit classes in the synthetic dataset Biased MNIST.}
  \label{tab:rgbvaluesBiasedMNIST}
\end{table}

\begin{table}[b]
  \centering
  \resizebox{\columnwidth}{!}{
  \begin{tabular}{@{}ccc@{}}
    \toprule
    Digit class & RGB values of left color & RGB values of right color\\
    \midrule
    0 & (250, 79, 42) & (4, 175, 212) \\
    1 & (252, 233, 89) & (2, 21, 165) \\
    2 & (171, 117, 147) & (83, 137, 107) \\
    3 & (199, 212, 153) & (55, 42, 101) \\
    4 & (22, 198, 250) & (232, 56, 4) \\
    5 & (81, 245, 113) & (173, 9, 141) \\
    6 & (6, 60, 193) & (248, 194, 61) \\
    7 & (141, 25, 194) & (113, 229, 60) \\
    8 & (52, 100, 4) & (202, 154, 250) \\
    9 & (212, 51, 68) & (42, 203, 186) \\
    \bottomrule
  \end{tabular}
  }
  \caption{RGB values of the colors associated with the digit classes in the synthetic dataset Multi-Color MNIST.}
  \label{tab:rgbvaluesMulticolorMNIST}
\end{table}

We present here a more detailed description of the datasets employed for our experiments.\\
\noindent\textbf{Biased MNIST.} Biased MNIST is a synthetic variant of the MNIST handwritten digit dataset, originally introduced in \cite{ReBias}, and widely used in the debiasing literature to evaluate the effectiveness of debiasing methods. The dataset is constructed by associating a specific background color to each of the ten digit classes (\eg 0: red, 1: green, etc.; see \cref{tab:rgbvaluesBiasedMNIST}). For a proportion $\rhodataset$ of the training samples, the background color matches the one assigned to the digit’s class. For the remaining $~{1-\rhodataset}$, a background color corresponding to a different class (selected uniformly at random) is applied. We report the RGB values of the colors associated with the digits in Biased MNIST in \cref{tab:rgbvaluesBiasedMNIST}.\\
\noindent\textbf{Multi-Color MNIST.} Multi-Color MNIST is yet another synthetic variant of the MNIST handwritten dataset, originally introduced in \cite{multicolor_mnist} and widely used in the debiaising literature to evaluate the effectiveness of debiaising methods on multiple biases. The dataset is constructed by associating two specific background color to the ten digit classes (\eg 0: left color is red, right color is light blue, etc.; see \cref{tab:rgbvaluesMulticolorMNIST}). For a proportion $\leftrhodataset$ (respectively $\rightrhodataset$) of the training samples, the left (respectively right) background color matches the one assigned to the digit's class. For the remaining $~{1-\leftrhodataset}$ (resp. $1-\rightrhodataset)$, a left (resp. right) background color corresponding to a different class (selected uniformly at random) is applied. $\leftrhodataset$ and $\rightrhodataset$ are fully tunable, thus allowing to experiment with different noise level combinations. We report the RGB values of the colors associated with the digits in Multi-Color MNIST in \cref{tab:rgbvaluesMulticolorMNIST}.\\
\noindent\textbf{BFFHQ.} Biased Flickr-Faces-HQ (BFFHQ) is a dataset of face images that builds upon FFHQ \cite{ffhq} by introducing demographic biases. The target of the generation is the age of the individual in the image, defined as $~{\mathcal{Y} = \{\text{``\texttt{young}"}, \text{``\texttt{old}"}\}}$, while the bias attribute is gender, defined as $~{\mathcal{B} = \{\text{``\texttt{female}"}, \text{``\texttt{male}"}\}}$. A proportion $~{\rhodataset=0.95}$ of images labeled as ``\texttt{young}" depict female subjects, while the same proportion of images labeled as ``\texttt{old}" depict male subjects.\\
\noindent\textbf{LAION-5B.} LAION-5B \cite{laion_5b} is a large-scale dataset consisting of image–captions pairs collected from the web. Due to its web-scraped nature, it inherently reflects a wide range of societal stereotypes \cite{uncurated_image_text_datasets_shedding_light_on_demographic_bias, amplify_demographic_bias}. We focus on the gender as a bias attribute ($~{\mathcal{B} = \{\text{``\texttt{female}"}, \text{``\texttt{male}"}\}}$), as it is widely studied in the literature \cite{survey_TTI_bias_evaluation}. The target concept for generation is the occupation ``\texttt{lawyer}". 
Unlike Biased MNIST and BFFHQ, estimating the baseline value $\rhodataset$ in LAION-5B is non-trivial. As highlighted in \cite{bias_amplification_paradox}, there exists a significant distributional shift between the captions used during model training and the prompts used at inference time for bias evaluation. Accurate estimation of $\rhodataset$ would require replicating the complex methodology proposed in \cite{bias_amplification_paradox}, which we omit for the sake of simplicity. Consequently, in the context of LAION-5B, we will focus on whether $\rhomodel$ changes when the sampling hyperparameters do, and not on the bias amplification phenomenon.\\

\section{Experimental details}
\label{sec:unets_architecture}

\begin{table}
  \centering
   \resizebox{\columnwidth}{!}{
  \begin{tabular}{@{}ccc@{}}
    \toprule
    Parameter & Biased MNIST & BFFHQ\\
    \midrule
    Resolutions & 32-16-8 & 64-32-16-8\\
    Residual blocks per resolution & 2 & 4\\
    Resolutions with attention & 16 & 16 \\
    Channels per resolution & 128-128-128 & 128-256-256-256\\
    Attention heads & 1 & 1  \\
    Attention blocks in encoder & 4 & 4\\
    Attention blocks in decoder & 2 & 2\\
    Nb trainable parameters & $8,797,955$ & $61,804,931$ \\
    \bottomrule
  \end{tabular}
  }
  \caption{Details of the architecture of the UNets used on the datasets Biased MNIST and BFFHQ.}
  \label{tab:unetsArchitecture}
\end{table}
\begin{figure*}
\begin{subfigure}{0.32\textwidth}
    \centering
    \includegraphics[width=1\linewidth]{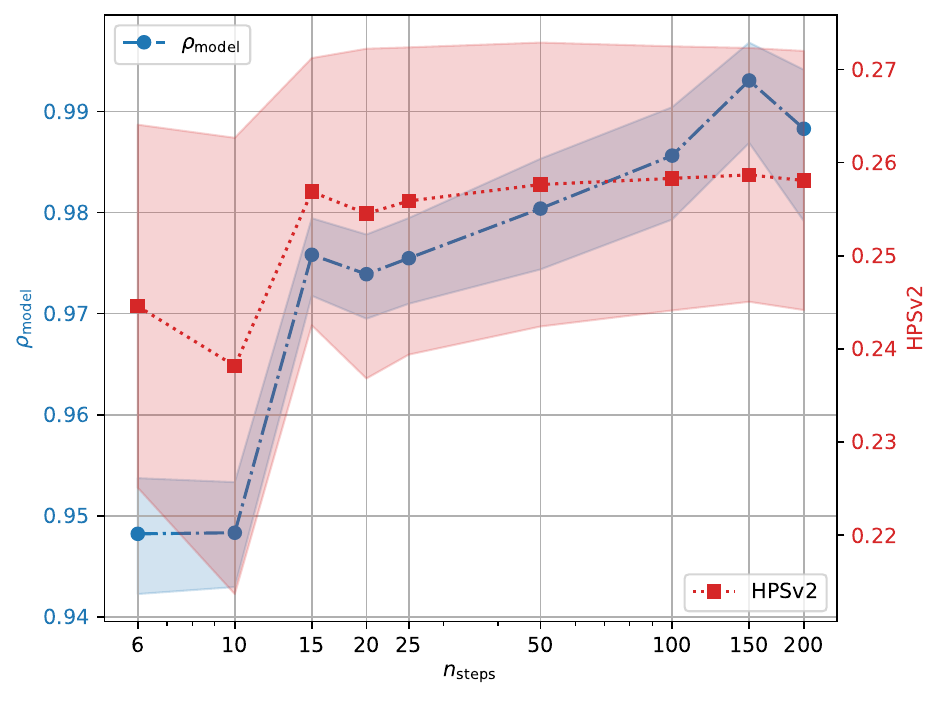}
    \caption{$\rhomodel$ and HPSv2 vs $\nsteps$ with $\eta\smeq 0.3$.}
    \label{fig:SD_portraitlawyer_rho_hps_eta03_logscale}
\end{subfigure}
\begin{subfigure}{0.32\textwidth}
    \centering
    \includegraphics[width=1\linewidth]{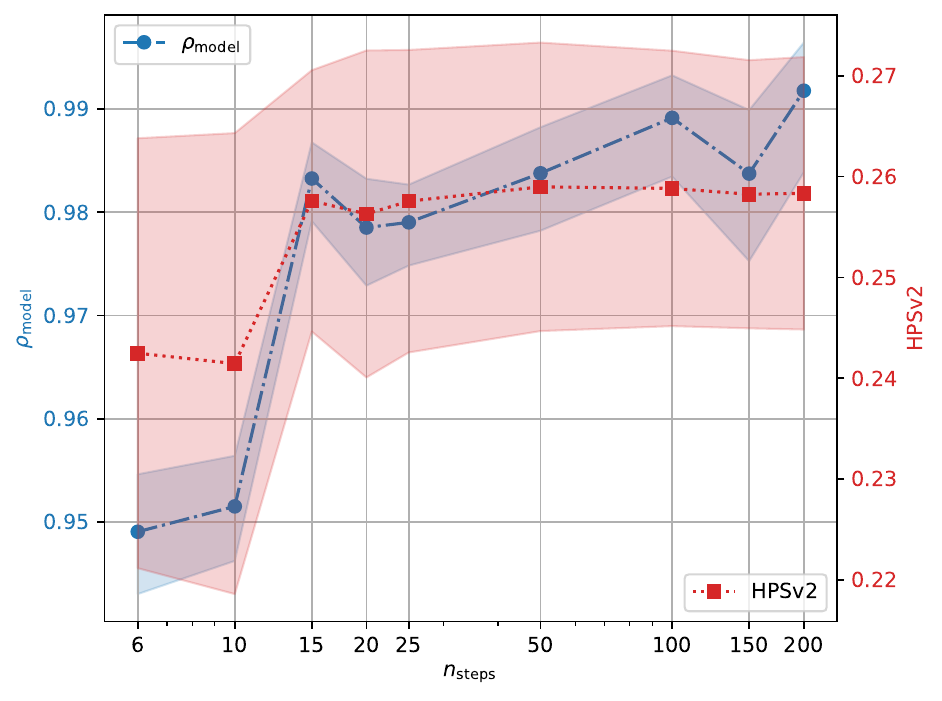}
    \caption{$\rhomodel$ and HPSv2 vs $\nsteps$ with $\eta\smeq 0.7$.}
    \label{fig:SD_portraitlawyer_rho_hps_eta07_logscale}
\end{subfigure}
\begin{subfigure}{0.32\textwidth}
    \centering
    \includegraphics[width=1\linewidth]{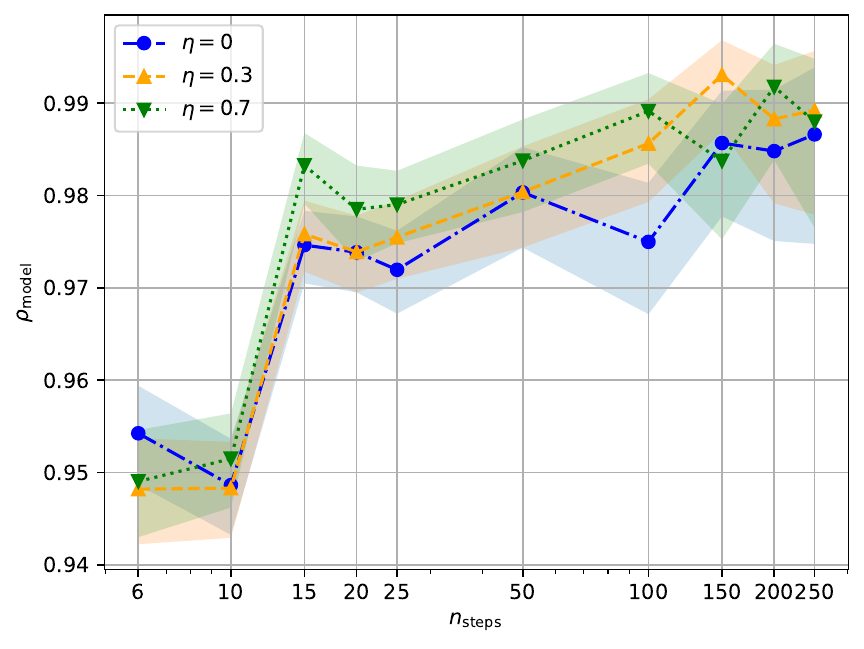}
    \caption{$\rhomodel$ vs $\nsteps$.}
    \label{fig:SD_portraitlawyer_eta0_0.3_0.7_varnsteps}
\end{subfigure}
\caption{Stable Diffusion v2.1 using DDIM sampler with $\eta\in\{0,0.3,0.7\}$ (deterministic sampling). See \cref{tab:numSamplesSD} for the number of samples used to compute each point. Note that the x-axis is on a logarithmic scale.}
\end{figure*}
\textbf{U-Net architecture.} In \cref{tab:unetsArchitecture} we report the main details of the network architectures used in this paper. We implemented them in a newly written codebase based loosely on the implementation by Song~\etal~\footnote{\href{https://github.com/yang-song/score_sde}{https://github.com/yang-song/score\_sde}}\cite{song_diffusion_sde} and based on the model and sampler implementation of Karras~\etal~\cite{edm}.\footnote{\href{https://github.com/NVlabs/edm}{https://github.com/NVlabs/edm}}\\
\textbf{Image generation.} The sampler we use on models trained on Biased MNIST, Multi-Color MNIST and BFFHQ is the stochastic sampler of Karras~\etal \cite{edm} (we also test VP-sampler \cite{song_diffusion_sde} and DPM-Solver \cite{dpm_solver} in the supplementary material). We make it vary by changing the hyperparameters $~{\{\nsteps, \Schurn, \Stmin, \Stmax, w\}}$ and by either using or not the second order correction. Regarding Stable Diffusion, we prompt it with ``A portrait photo of a lawyer". We use the DDIM sampler, which allows us to control the number of sampling steps with $\nsteps$ and the stochasticity with the parameter $\eta$.  $~{\eta=0}$ corresponds to deterministic sampling and $~{\eta=1}$ introduces as much variance in the process as in the ancestral sampling of DDPM~\cite{DDPM}.\\
\textbf{Image selection.} We keep all the images generated by models trained on Biased MNIST. We only keep an image generated by Stable Diffusion or the model trained on BFFHQ if the face of the individual in the image is clearly visible. We use the OpenCV 8-bit quantized version of the Single-Shot-Multibox face detector to detect the face. 
For the model trained on BFFHQ, we only keep the image if the face detected with the highest confidence has a confidence level above $0.999$. For Stable Diffusion, the image is kept if: a single face is detected, the confidence is above $0.95$, and the bounding box is at least 10 pixels away from all borders.\\
\textbf{Image quality assessment.} We evaluate the quality of the generated images for Stable Diffusion using the Human Preference Score v2 (HPSv2)~\cite{hpsv2}, instead of the FID metric \cite{fid_metrics} used for Biased MNIST and BFFHQ. HPSv2 presents two key advantages over the FID: it eliminates the need for training samples with captions resembling the prompt and exhibits a higher correlation with human preferences. HPSv2 is trained by finetuning CLIP \cite{clip_score} 
on version 2 of the Human Preference Dataset~\cite{hpsv2}.\\
\begin{table}
  \centering
  \resizebox{0.5\columnwidth}{!}{
  \begin{tabular}{@{}cccc@{}}
    \toprule
    $\nsteps$ & $\eta=0$ & $\eta=0.3$ & $\eta=0.7$\\
    \midrule
    6 & 5990 & 5908 & 5691 \\
    10 & 7127 & 7178 & 7031 \\
    15 & 6427 & 6412 & 4603 \\
    20 & 6045 & 5863 & 3212 \\
    25 & 5422 & 5344 & 5387 \\
    50 & 2697 & 2653 & 2651 \\
    100 & 1999 & 1950 & 1935 \\
    150 & 1328 & 1300 & 1292 \\
    200 & 988 & 940 & 973 \\
    250 & 673 & 650 & 667 \\
    \bottomrule
  \end{tabular}
  }
  \caption{Number of samples generated by Stable Diffusion used to compute $\rhomodel$ for every $\nsteps$ and every $\eta$.}
  \label{tab:numSamplesSD}
\end{table}
\noindent\textbf{Stable Diffusion image count.} In \cref{tab:numSamplesSD}, we report the number of images used to compute every point of Fig.~6a, \cref{fig:SD_portraitlawyer_rho_hps_eta03_logscale}, \cref{fig:SD_portraitlawyer_rho_hps_eta07_logscale}, and \cref{fig:SD_portraitlawyer_eta0_0.3_0.7_varnsteps}. They vary for two reasons. First, we did not generate the same number of samples for every $\nsteps$ because the computational cost is linear in $\nsteps$, and therefore we could generate more images for lower $\nsteps$. Then, among the generated images, we had to remove those where the face was not clearly visible, following the process described in Sec.~4.1, which caused the number of remaining images to vary between the different $\eta$.\\
\textbf{More details on results in the main paper.} Here we summarize the specific configurations chosen to display the results in the main paper:
\begin{itemize}
\item Fig.~4: $\{\nsteps\smeq25, \Schurn\smeq80, \Stmin\smeq0.01, \Stmax\smeq80\}$. Each point is obtained by using the estimator in Eq.~(8) on 4000 generated images. The unconditional score necessary for CFG is computed by averaging the conditional scores (as in Eq.~(6)).
\item Fig.~6a: $\{\Schurn\smeq 80, \Stmax\smeq 80\}$.
\item Fig.~6b: $\{\Schurn\smeq 80, \Stmin\smeq 0.1\}$.
\item Fig.~7: $\{\Stmin\smeq 0.1, \Stmax\smeq 80\}$.
\end{itemize}
Please note that the same effects described are observable for a broad set of hyperparameter choices - they are here chosen for visualization purposes only.

\begin{figure*}[t]
    \centering
    \includegraphics[width=1\linewidth]{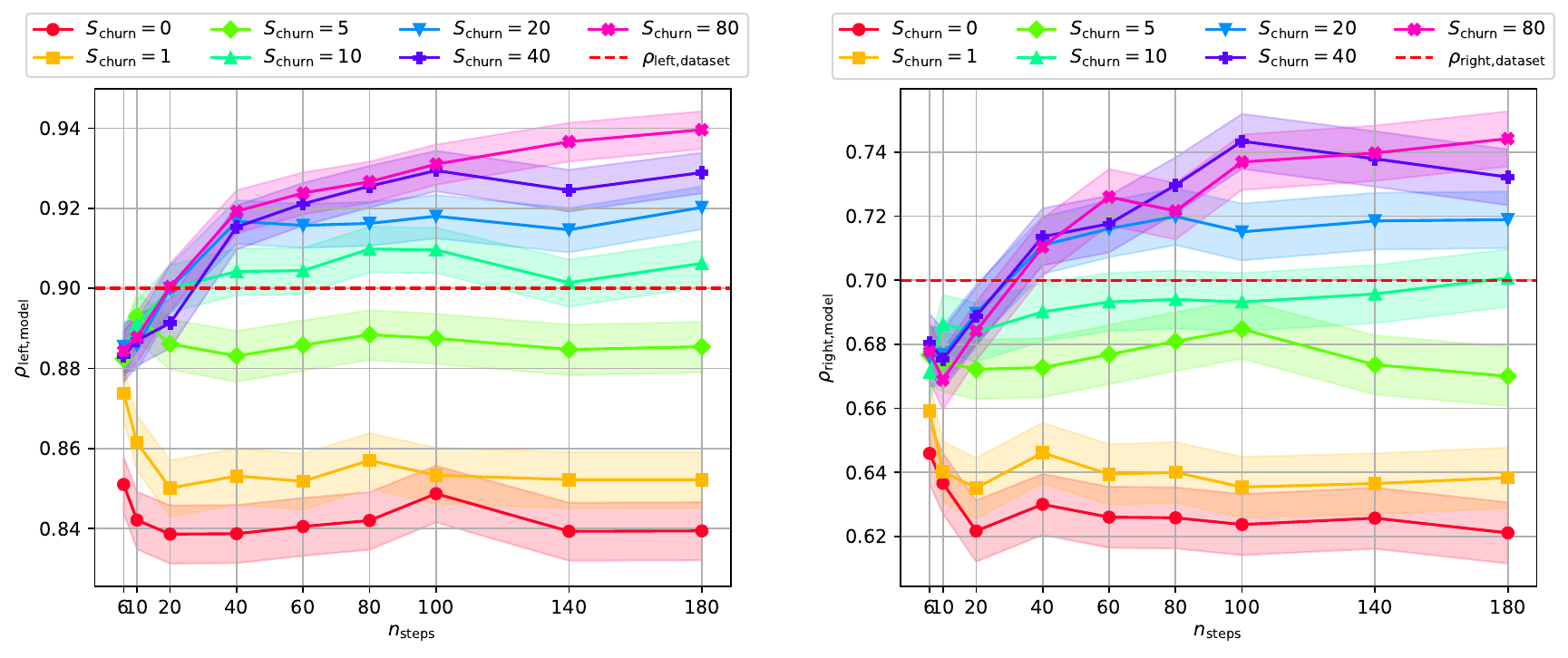}
    \caption{$\leftrhomodel$ and $\rightrhomodel$ vs $\nsteps$ for various $\Schurn$ for a model trained on Multi-Color MNIST ($\leftrhodataset\smeq 0.9$, $\rightrhodataset\smeq 0.7$) using EDM sampler (same experiment as \cref{fig:mcmnist_leftrho=0.9_rightrho=0.7_edmsampler_varSchurn}).}
    \label{fig:mcmnist_leftrho=0.9_rightrho=0.7_edmsampler_varnsteps}
\end{figure*}
\begin{figure*}[t]
    \centering
    \includegraphics[width=1\linewidth]{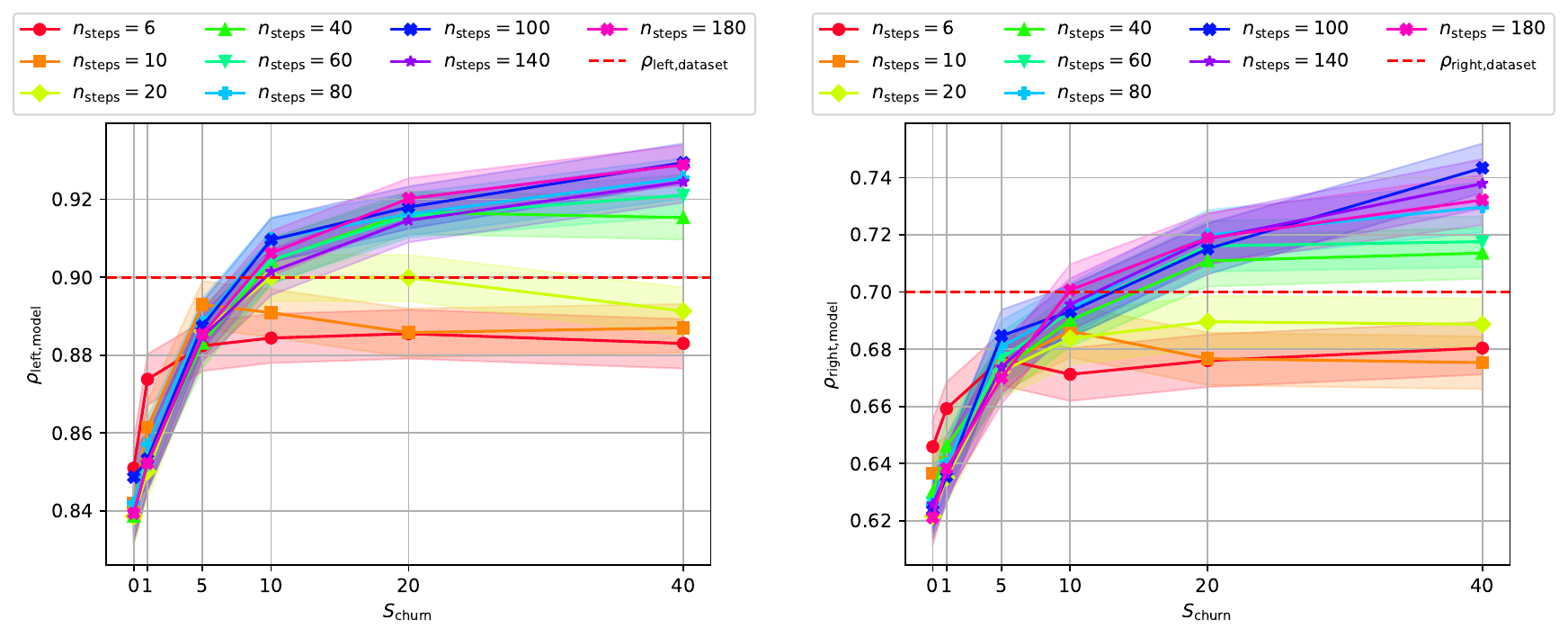}
    \caption{$\leftrhomodel$ and $\rightrhomodel$ vs $\Schurn$ for various $\nsteps$ for a model trained on Multi-Color MNIST ($\leftrhodataset\smeq 0.9$, $\rightrhodataset\smeq 0.7$) using EDM sampler (same experiment as \cref{fig:mcmnist_leftrho=0.9_rightrho=0.7_edmsampler_varnsteps}).}
    \label{fig:mcmnist_leftrho=0.9_rightrho=0.7_edmsampler_varSchurn}
\end{figure*}
\section{More experimental results}
\label{sec:additional_exp_results}
\label{sec:addit_res_guidance_scale}

\begin{figure*}
    \centering
    \includegraphics[width=1\linewidth]{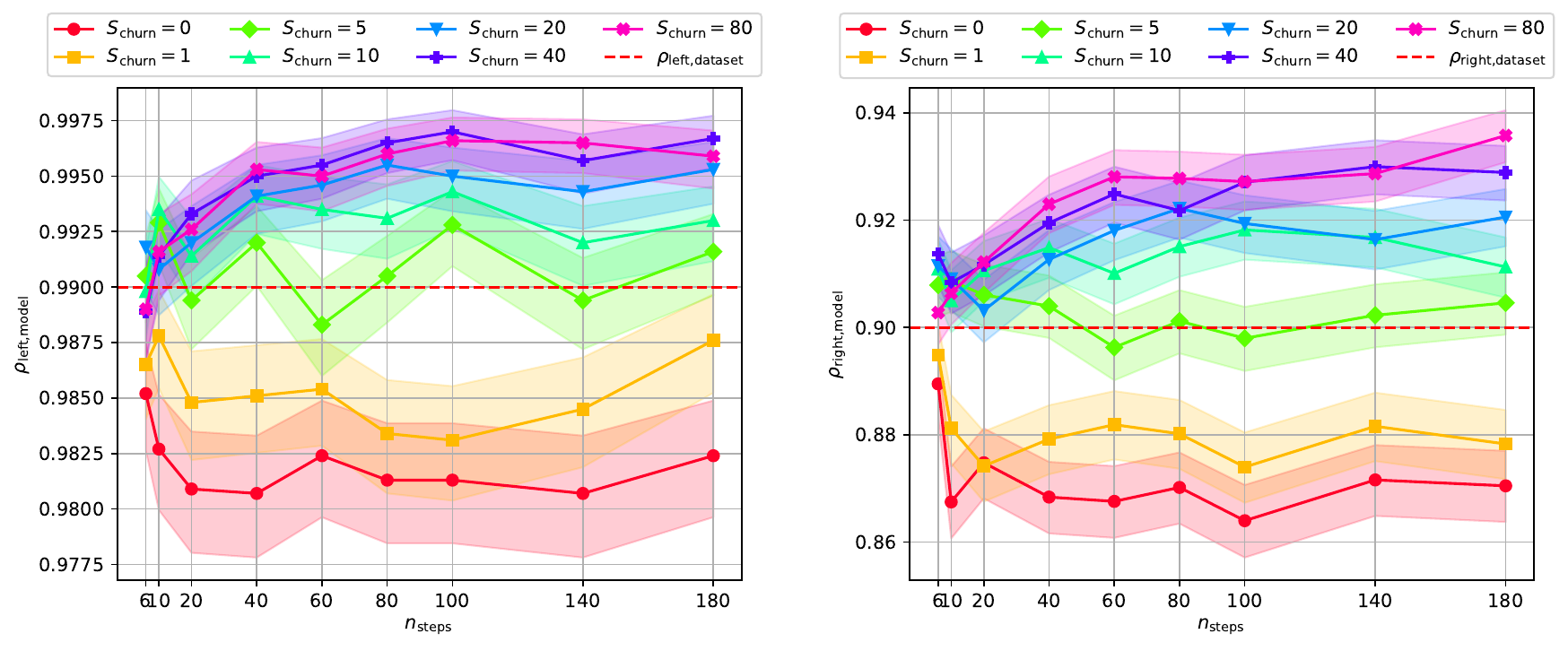}
    \caption{$\leftrhomodel$ and $\rightrhomodel$ vs $\nsteps$ for various $\Schurn$ for a model trained on Multi-Color MNIST ($\leftrhodataset\smeq 0.99$, $\rightrhodataset\smeq 0.9$) using EDM sampler (same experiment as \cref{fig:mcmnist_leftrho=0.99_rightrho=0.9_edmsampler_varSchurn}).}
    \label{fig:mcmnist_leftrho=0.99_rightrho=0.9_edmsampler_varnsteps}
\end{figure*}

\begin{figure*}
    \centering
    \includegraphics[width=1\linewidth]{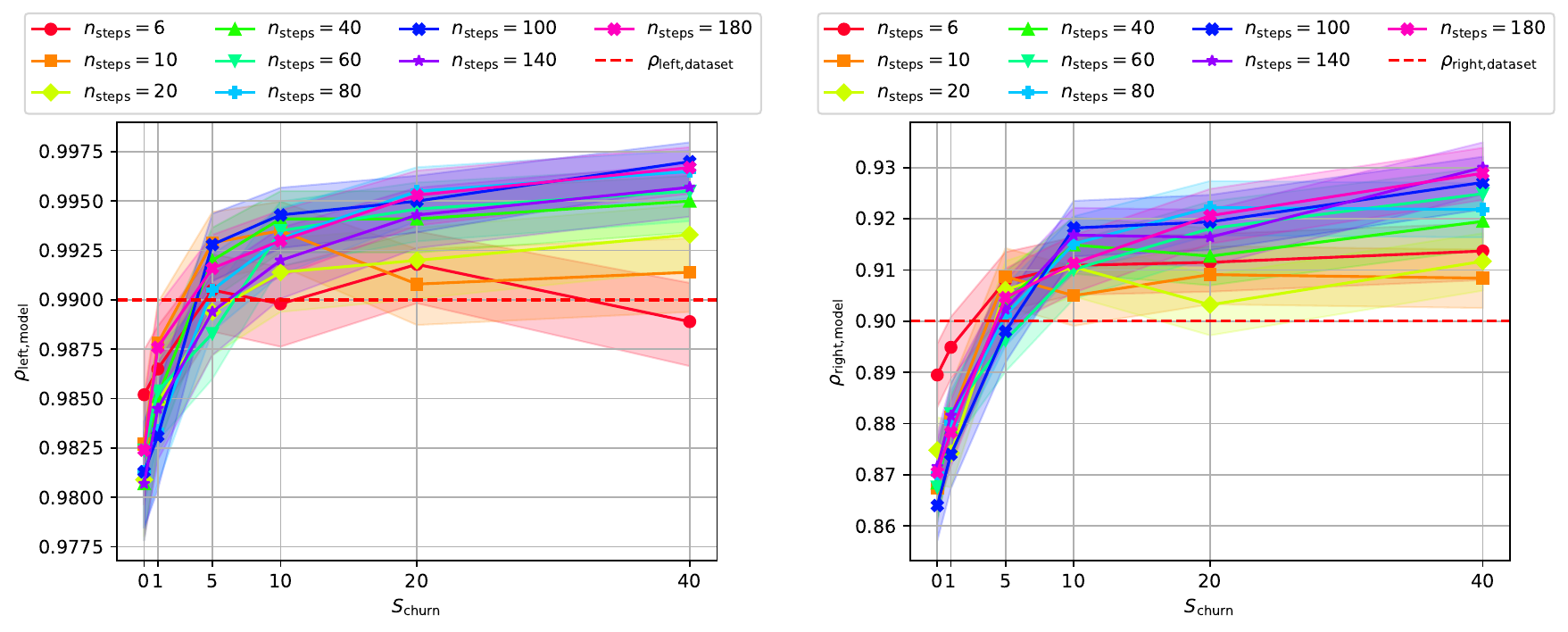}
    \caption{$\leftrhomodel$ and $\rightrhomodel$ vs $\Schurn$ for various $\nsteps$ for a model trained on Multi-Color MNIST ($\leftrhodataset\smeq 0.99$, $\rightrhodataset\smeq 0.9$) using EDM sampler (same experiment as \cref{fig:mcmnist_leftrho=0.99_rightrho=0.9_edmsampler_varnsteps}).}
    \label{fig:mcmnist_leftrho=0.99_rightrho=0.9_edmsampler_varSchurn}
\end{figure*}

\begin{figure*}
\centering
\begin{subfigure}{0.49\textwidth}
    \centering
    \includegraphics[width=1\linewidth]{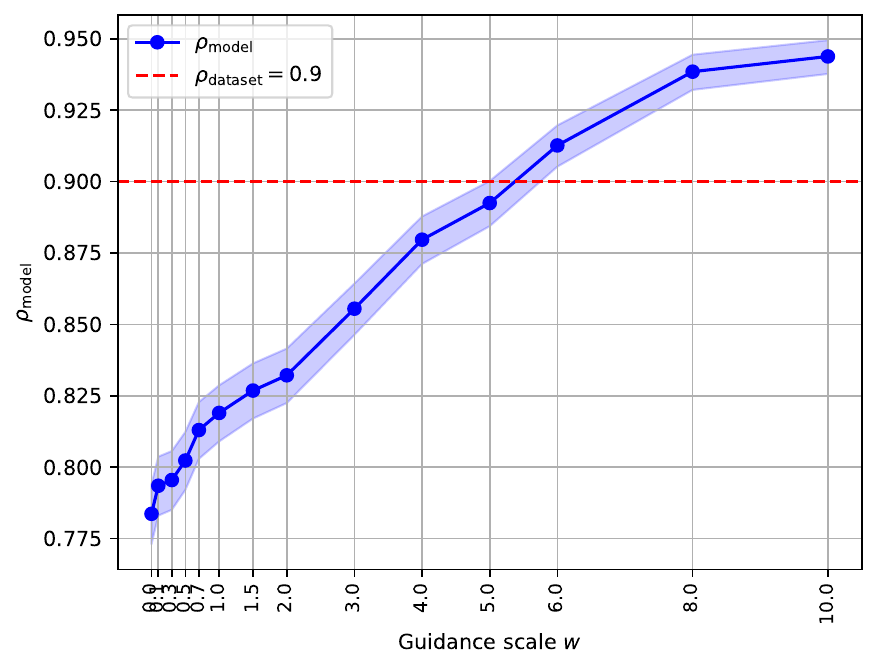}
    \caption{$\rhomodel$ vs $w$ for a model trained on 2-classes Biased MNIST ($\rhodataset\smeq0.9)$) using DPM-Sovler-1 with hyperparameter $\{\nsteps=10\}$.}
    \label{fig:2clbmnist_0.9_dpmsolver1_var_guidance_scale_nsteps=10}
\end{subfigure}
~
\begin{subfigure}{0.49\textwidth}
    \centering
    \includegraphics[width=1\linewidth]{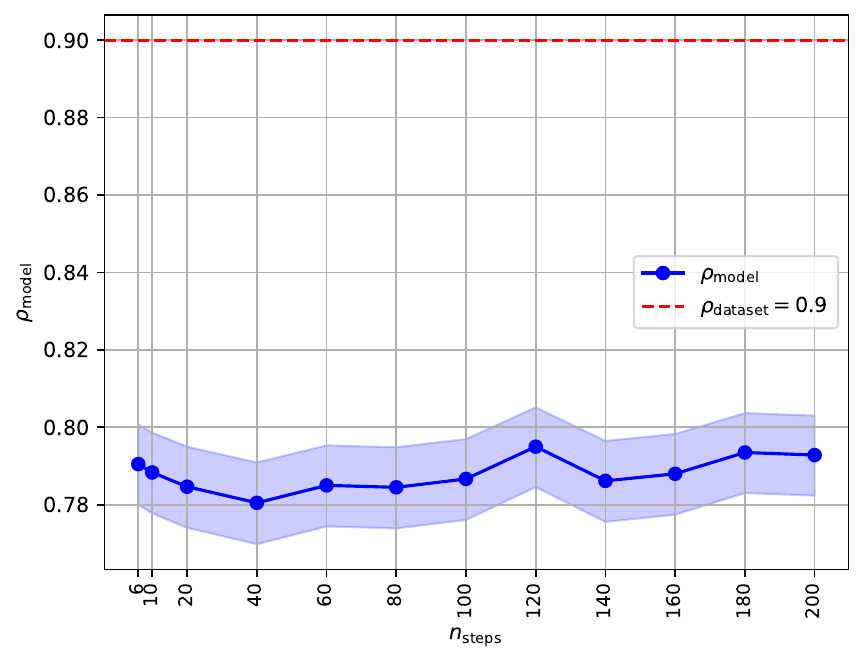}
    \caption{$\rhomodel$ vs number of times teps on 2-classes Biased MNIST ($\rhodataset\smeq0.9$) using DPM-Sovler-1 with hyperparameter $\{w=0\}$.}
    \label{fig:2clbmnist_0.9_dpmsolver1_var_nsteps_w=0}
\end{subfigure}
\caption{DPM-Solver-1 results on 2-classes Biased MNIST.}
\end{figure*}

\begin{figure}
    \centering
    \includegraphics[width=1\linewidth]{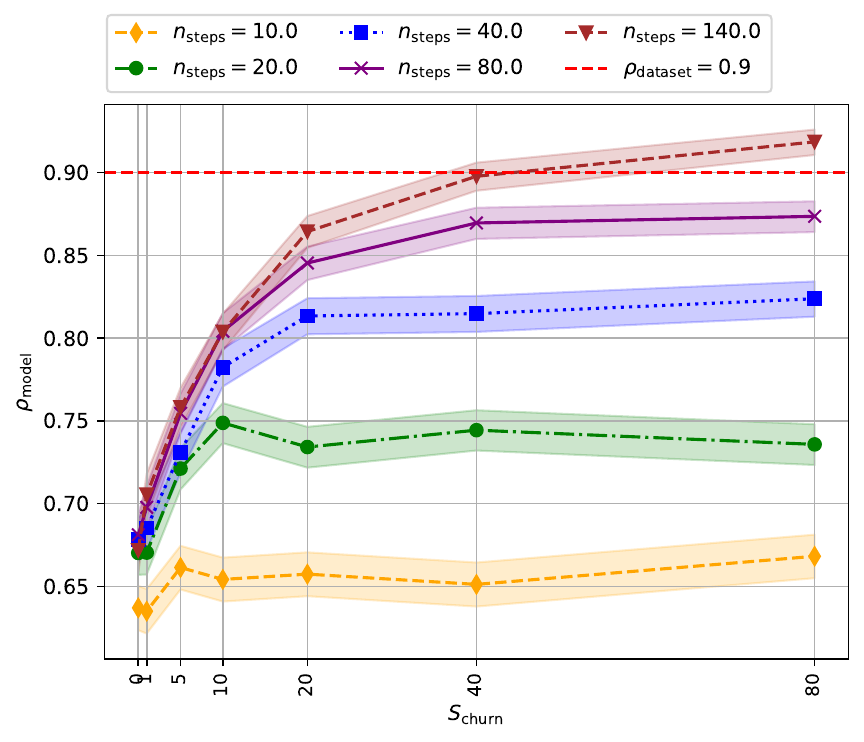}
    \caption{$\rhomodel$ vs $\Schurn$ for various $\nsteps$ for a model trained
on 10-classes Biased MNIST ($\rhodataset$=0.9) using VP sampler.}
    \label{fig:10clbmnist_0.9_vpsolver_var_nsteps_var_s_churn}
\end{figure}

\begin{figure*}
    \centering
    \includegraphics[width=0.65\linewidth]{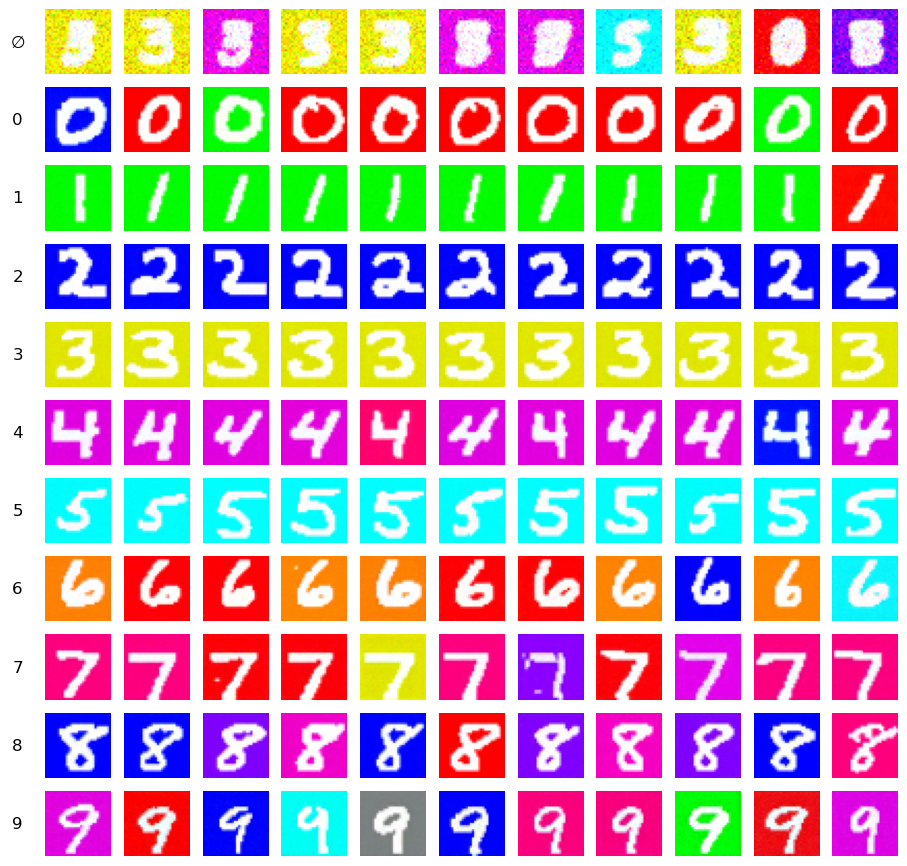}
    \caption{Samples generated by model trained on 10-classes Biased MNIST ($\rhodataset\smeq0.9$) using Karras deterministic sampler ($\Schurn \smeq0, \nsteps\smeq12$) with guidance scale $w\smeq12$. Each row is sampled with a different conditioning: the first row are unconditional results, the others are obtained by conditioning on the class written at the beginning of the row. The unconditional score prediction is obtained by averaging the score prediction over all classes (see Eq.~(6)).}
    \label{fig:10cl_bmnist_guidscale12_nsteps12_schurn0_aggregscore_samples}
\end{figure*}

\begin{figure*}
    \centering
    \includegraphics[width=0.65\linewidth]{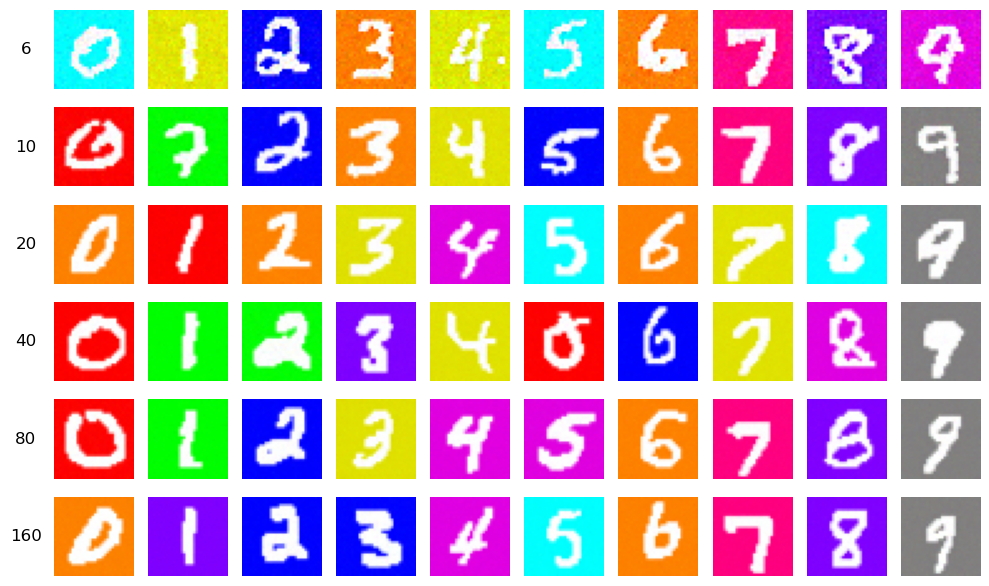}
    \caption{Samples obtained with a model trained on Biased MNIST ($\rhodataset\smeq 0.7$) using Karras stochastic sampler with hyperparameters $\{\Schurn\smeq60, \Stmin\smeq0.01, \Stmax\smeq80\}$. The number at the beginning of each row indicates the number of sampling steps for the row. Each column corresponds to a single class. All images in the same column were generated starting from the same initial noise. Except for the samples generated with $\nsteps\smeq 6$, there is no noticeable quality difference between the different $\nsteps$.} 
    \label{fig:10cl_rho0.7_samples_varnsteps_samexinit}
\end{figure*}

\begin{figure*}
\centering
\begin{subfigure}{0.49\textwidth}
    \centering
    \includegraphics[width=1\linewidth]{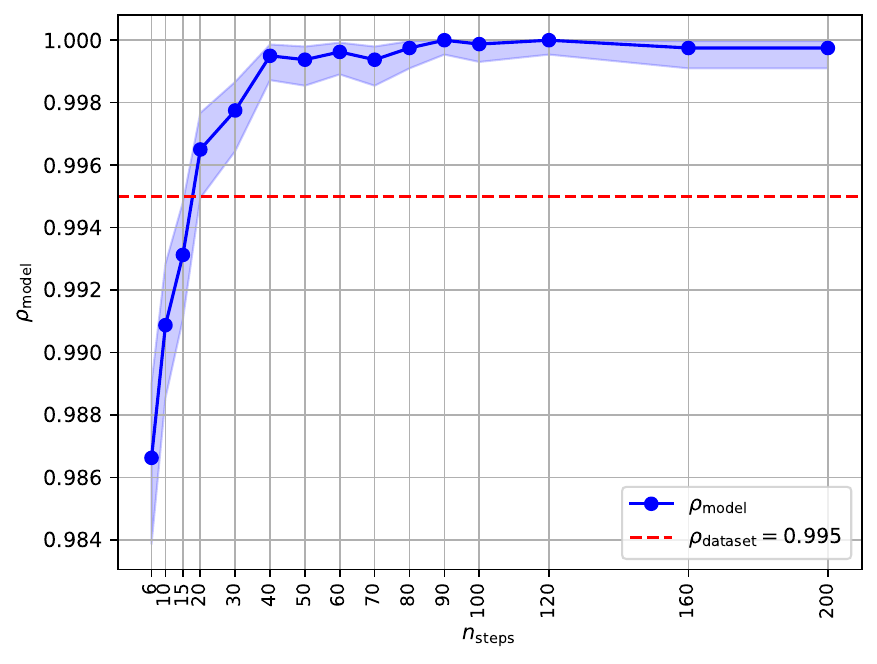}
    \caption{$\rhomodel$ vs $\nsteps$ for a model trained on 2-classes Biased MNIST ($\rhodataset\smeq 0.995$) using Karras stochastic sampler with hyperparameters $\{\Schurn\smeq40, \Stmin\smeq0.05, \Stmax\smeq50\}$. Each point is obtained by using the estimator as in Eq.~(8) on 8000 generated images.}
    \label{fig:2clbmnist_0.995_varnsteps_smin0.05_smax50_schurn40}
\end{subfigure}
~
\begin{subfigure}{0.49\textwidth}
    \centering
    \includegraphics[width=1\linewidth]{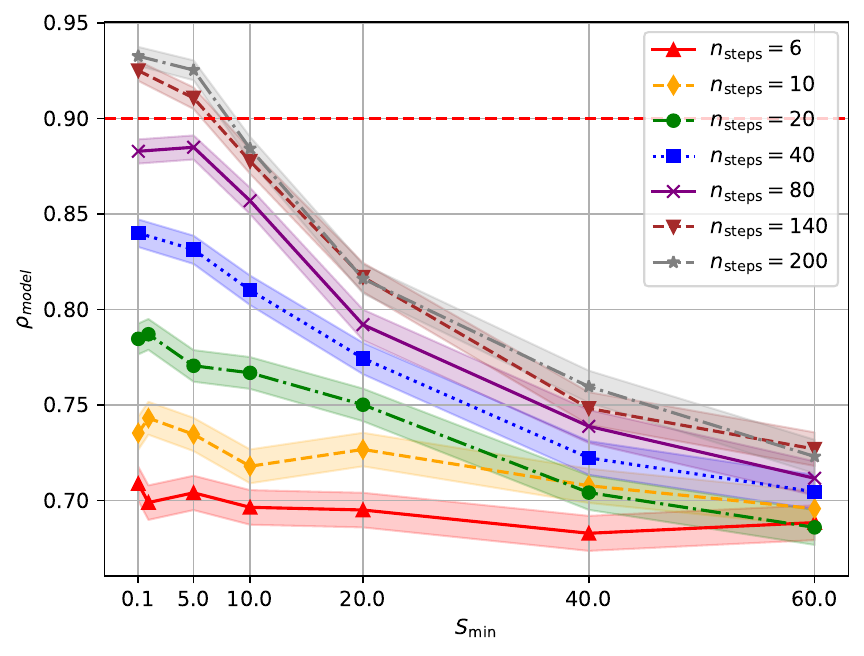}
    \caption{$\rhomodel$ vs $\Stmin$ for a model trained on 10-classes Biased MNIST ($\rhodataset\smeq 0.9$) using Karras stochastic sampler with fixed hyperparameters $\{\Schurn\smeq 80, \Stmax\smeq 80\}$ and varying $\nsteps$. Same data as in Fig.~7 but presented differently.}
    \label{fig:10clBmnist_rho0.9_vs_smin_schurn80_smax80_varnsteps}
\end{subfigure}
\caption{Supplementary results on Biased MNIST.}
\end{figure*}

\begin{figure*}
\centering
\begin{subfigure}{0.49\textwidth}
    \centering
    \includegraphics[width=1\linewidth]{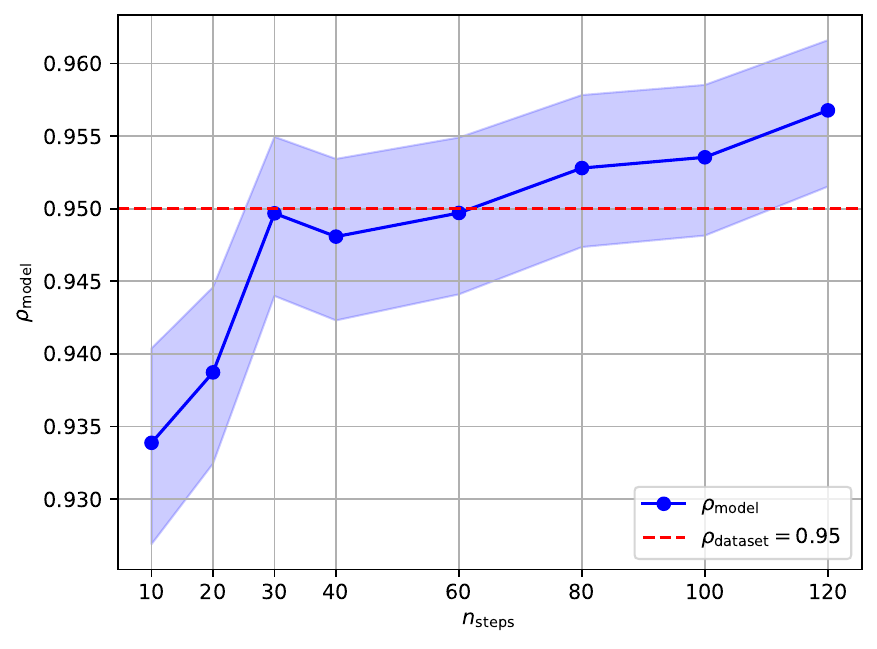}
    \caption{$\rhomodel$ vs $\nsteps$ for a model trained on BFFHQ ($\rhodataset\smeq0.95)$) using Karras stochastic samplers with hyperparameters $\{\Schurn\smeq80,\Stmin\smeq 0.01, \Stmax\smeq 80\}$.}
    \label{fig:BFFHQ_rho0.95_varNsteps_smin0.01_smax80_schurn80}
\end{subfigure}
~
\begin{subfigure}{0.49\textwidth}
    \centering
    \includegraphics[width=1\linewidth]{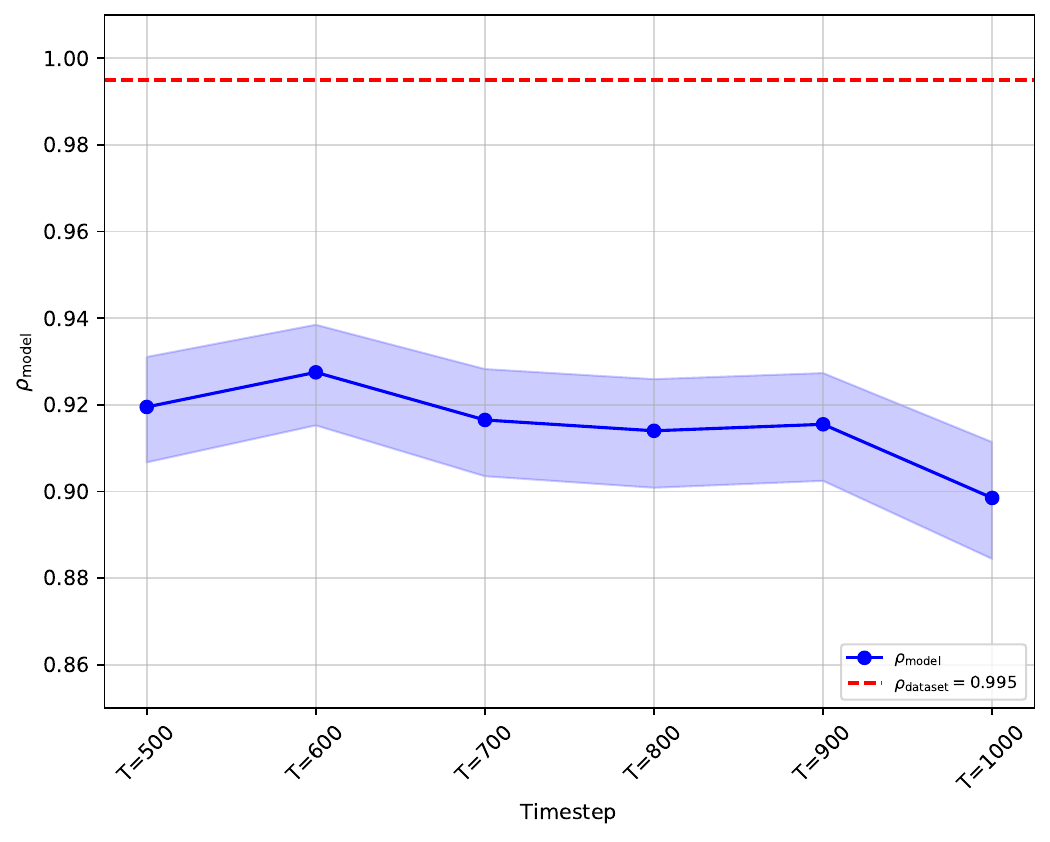}
    \caption{$\rhomodel$ vs number of time steps on BFFHQ ($\rhodataset\smeq0.995$) using classic DDPM version. Each point is obtained by using the oracle to classify the generated images and the bias.}
    \label{fig:var_timesteps_bffhq_ddpm}
\end{subfigure}
\caption{Supplementary results on BFFHQ.}
\end{figure*}

\begin{figure*}[t]
\centering
    \centering
    \includegraphics[width=1\linewidth]{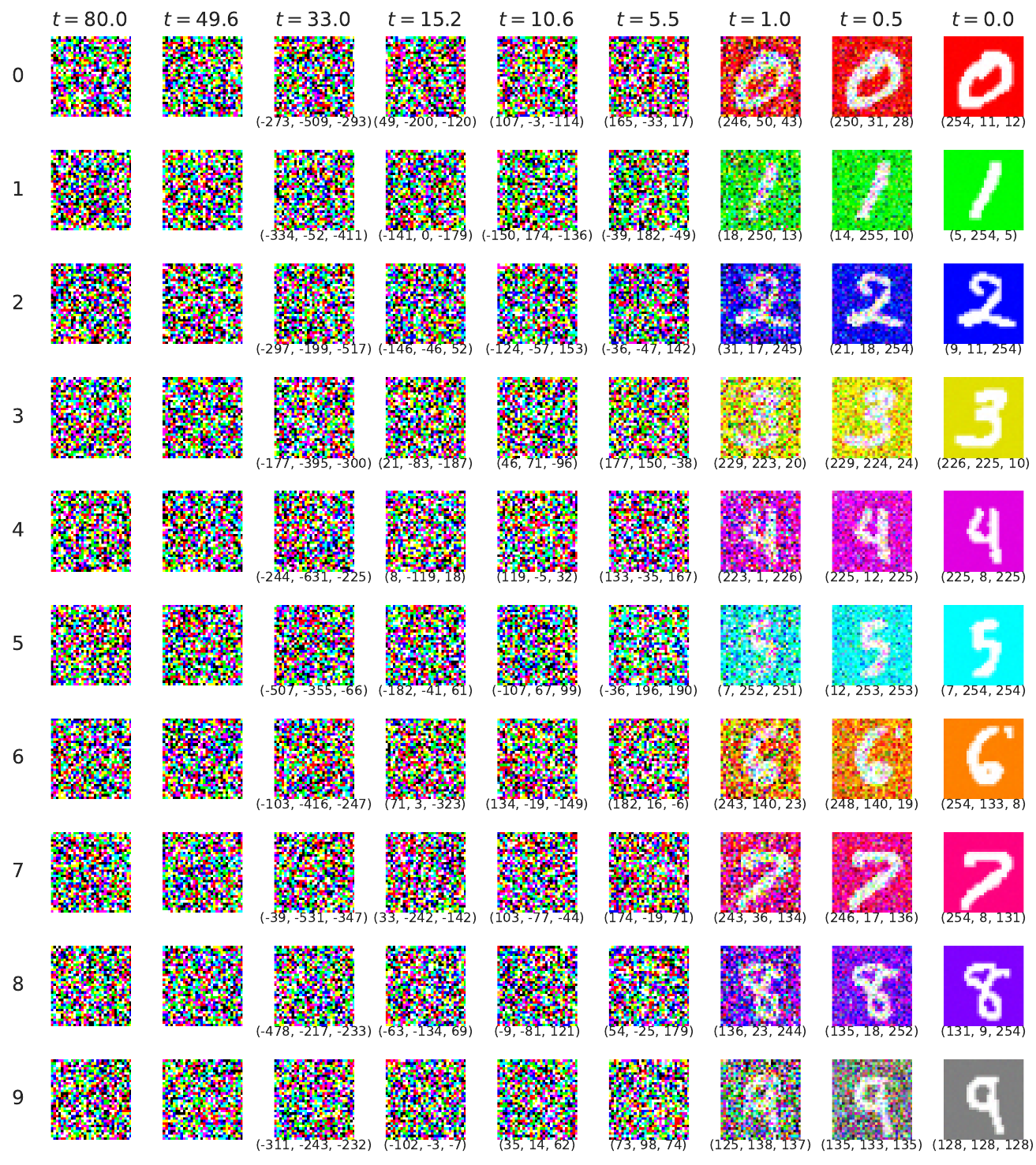}
    \caption{History of the generation of 10 samples (one per row) by a model trained on 10-classes Biased MNIST ($\rhodataset\smeq 0.9$) using Karras stochastic sampler with hyperparameters $\{\nsteps\smeq 60, \Schurn\smeq 80,\Stmin\smeq 0.01,\Stmax\smeq 80\}$. Each column corresponds to a single time step. Under the images is the RGB average of the non-white pixels (it can become negative for high noise values). Starting from $t=15.2$, it becomes possible to guess the end color of the image for the 6 first classes by setting the highest values to 255 and the others to 0. It means that even if the images remain very noisy, the bias has already started to appear.}
    \label{fig:10clBMINST_rho0.9_samples_history_noise}
\end{figure*}

\textbf{Dataset with multiple biases.}
Our previous experiments focused on datasets with a single bias attribute. We now verify our claims on Multi-Color MNIST \cite{multicolor_mnist}, a dataset with multiple known and controlled biases. In Multi-Color MNIST, the black background of each digit is split into two and filled with one color on the left and another color on the right. Similarly to Biased MNIST, the correlation between the colors and the digit is controlled by $\leftrhodataset$ and $\rightrhodataset$. Although the data set is quite simple, it is still relevant because the bias levels are fully tunable and we can assess $\rhomodel$ (left and right) with perfect accuracy. We present two combinations of the level of bias: one with medium bias $\{\leftrhodataset=0.9, \rightrhodataset=0.7\}$ and one with high bias $\{\leftrhodataset=0.99, \rightrhodataset=0.9\}$. In  \cref{fig:mcmnist_leftrho=0.9_rightrho=0.7_edmsampler_varnsteps} and \cref {fig:mcmnist_leftrho=0.9_rightrho=0.7_edmsampler_varSchurn} we present the results for medium bias, and in \cref{fig:mcmnist_leftrho=0.99_rightrho=0.9_edmsampler_varnsteps} and \cref{fig:mcmnist_leftrho=0.99_rightrho=0.9_edmsampler_varSchurn} the results for high bias. Our previous claims hold for $\leftrhomodel$ and $\rightrhomodel$ in both settings : $\nsteps$ and $\Schurn$ are positively correlated with $\leftrhomodel$ and $\rightrhomodel$. Moreover, for $\nsteps$ to affect the level of bias, a minimal amount of noise in the sampling ($\Schurn\approx10)$ is required.
Overall, the conclusion that we have drawn from a single bias attribute remains valid for multiple biases. 

\noindent\textbf{Additional samplers for continuous framework.} In addition to Karras' deterministic and stochastic samplers, we test two other samplers within the continuous framework. Specifically, we test the deterministic and stochastic VP-SDE sampler from Song \etal \cite{song_diffusion_sde} and the deterministic DPM-Solver-1 \cite{dpm_solver}.

The VP-SDE is a diffusion SDE introduced by Song \etal \cite{song_diffusion_sde} in their seminal paper. It has a different noise schedule and scaling from the SDE used in EDM \cite{edm}, resulting in a significantly different sampling trajectory. We refer to as the VP sampler the integration of the VP-SDE with the EDM scheme. The VP-SDE sampler has previously been studied and implemented by Karras \etal \cite{edm}. We measure $\rhomodel$ as incurred by the VP sampler on a model trained on 10-classes Biased MNIST using the same model and the same setup as in the experiment on the EDM sampler, with results presented in Figure 8.
Thus, we vary both $\Schurn$ and $\nsteps$. The results with the VP sampler in \cref{fig:10clbmnist_0.9_vpsolver_var_nsteps_var_s_churn} are extremely similar to those obtained with the EDM sampler, namely with a positive correlation between $\Schurn$ and $\rhomodel$ and between $\nsteps$ and $\rhomodel$, as well as the fact that the values of $\rhomodel$ obtained with different $\nsteps$ for $\Schurn\smeq 0$ are extremely close. This experiment shows that the results of our experiments on a given model carry over to a different sampler with different sampling trajectories, suggesting that the observed effects are a general phenomenon rather than the specifics of a particular sampler.

DPM-Solver is a deterministic method to efficiently sample from the diffusion models, leveraging the semi-linear structure of the probability flow ODE. In \cref{fig:2clbmnist_0.9_dpmsolver1_var_guidance_scale_nsteps=10} and \cref{fig:2clbmnist_0.9_dpmsolver1_var_nsteps_w=0}, we observe that the $\rhomodel$ measured on a model trained on 2-classes Biased MNIST with DPM-Sampler-1 is independent of the number of sampling steps and positively correlated with the guidance scale. These findings are not new, but they corroborate the observation that in small models the number of sampling steps affects the level of bias only when $\Schurn>0$, \ie when the sampling process is stochastic rather than deterministic.\\
\textbf{Effect of conditioning strength.}
We observe in \cref{fig:10cl_bmnist_guidscale12_nsteps12_schurn0_aggregscore_samples} that with a high guidance scale, the biases of the classes from 0 to 5 are well represented and amplified as expected, but the biases from classes 6 to 9 are under-represented, or even not represented at all in the case of 9. It turns out that the RGB values (see \cref{tab:rgbvaluesBiasedMNIST}) of the colors of classes 0 to 5 are in the corners of the cube $[0;255]^3$ in the RGB space, whereas those of classes 6 to 8 are in the middle of the edges, and that the RGB value of the color correlated with class 9 is in the center of the cube. Keeping in mind that the effect of CFG is to add guidance (the second term in Eq.~(5)) that ``pushes" the samples away from the mean (unconditional) data distribution and towards the class distribution, we interpret this result as the CFG pushing the colors of the samples away from the mean color distribution (which is gray, at the center of the RGB cube), and towards the corners. Thus, the nature of the bias in 10-class Biased MNIST makes it unfit to properly study the effect of $w$, so we resort to 2-classes Biased MNIST, where the colors of both classes (red and green) play a symmetrical role.
\begin{figure*}[t]
    \centering
    \begin{subfigure}[b]{0.49\textwidth}
    \centering
    \includegraphics[width=0.95\linewidth]{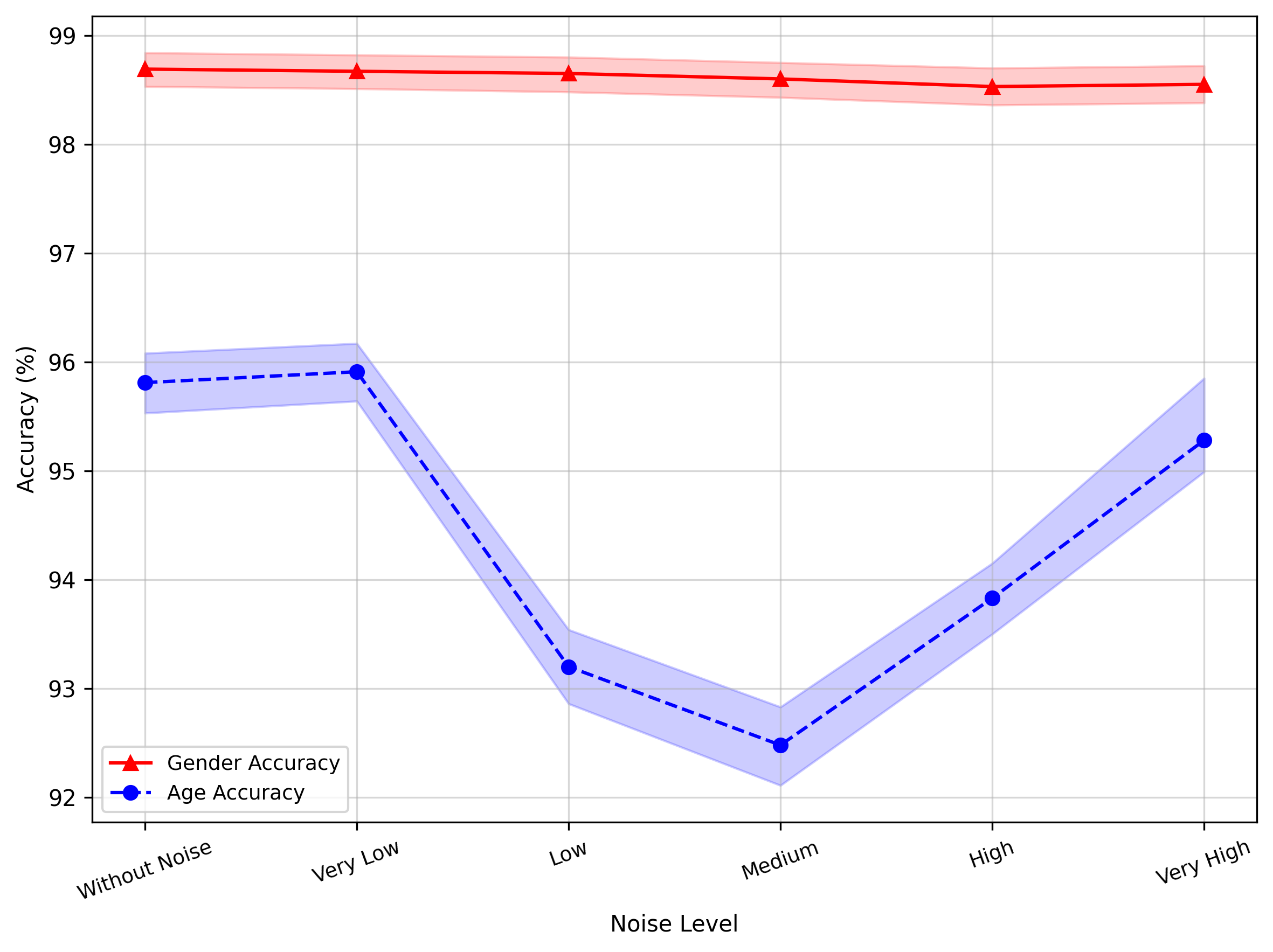}
    \caption{Accuracy of the oracle in gender and age prediction.}
    \label{fig:oracle_cdpm_accuracy_total}
    \end{subfigure}
    \begin{subfigure}[b]{0.49\textwidth}
    \centering
    \includegraphics[width=0.95\linewidth]{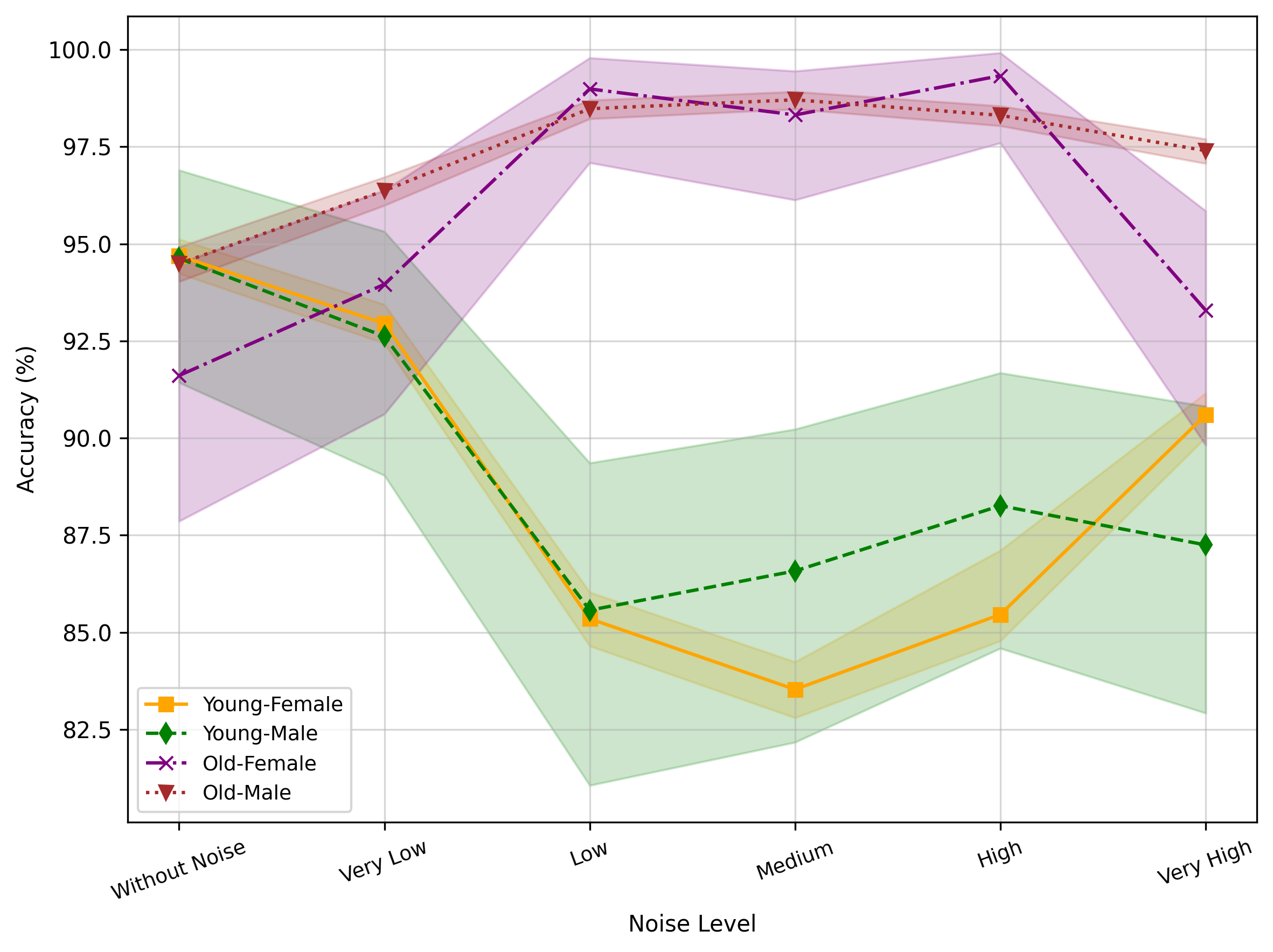}
    \caption{Group-wise accuracy of the oracle in gender and age prediction.}
    \label{fig:oracle_cdpm_accuracy_groups}
    \end{subfigure}
    \caption{Analysis of different levels of noise injected in the latent space of Stable Diffusion.}
\end{figure*}
\begin{figure*}
    \centering
    \includegraphics[width=0.95\linewidth]{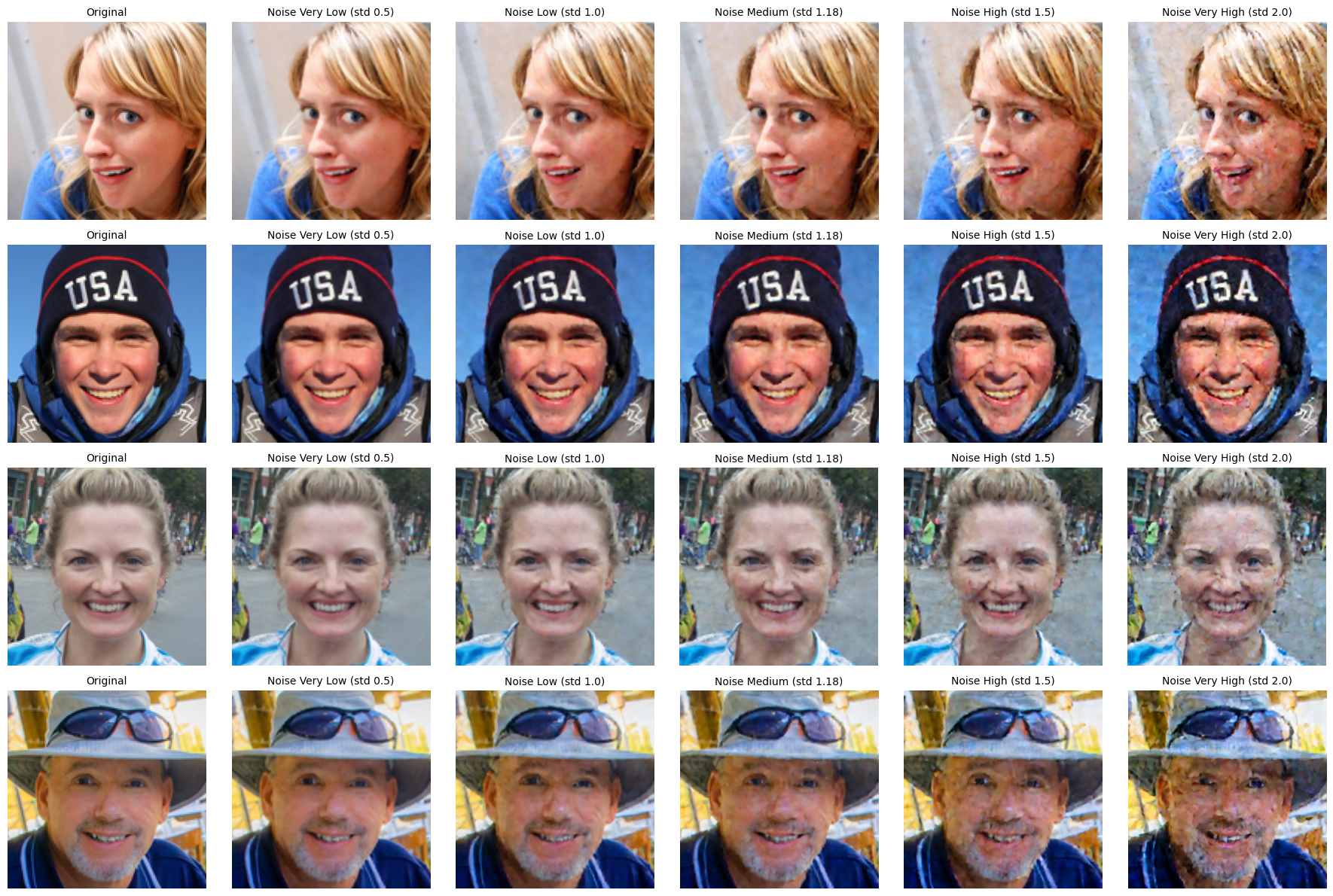}
    \caption{ Random train samples with all considered levels of noise injected in the latent space of Stable Diffusion.}
    \label{fig:noisy_train_samples}
\end{figure*}
\begin{figure*}
    \centering
    \includegraphics[width=0.95\linewidth]{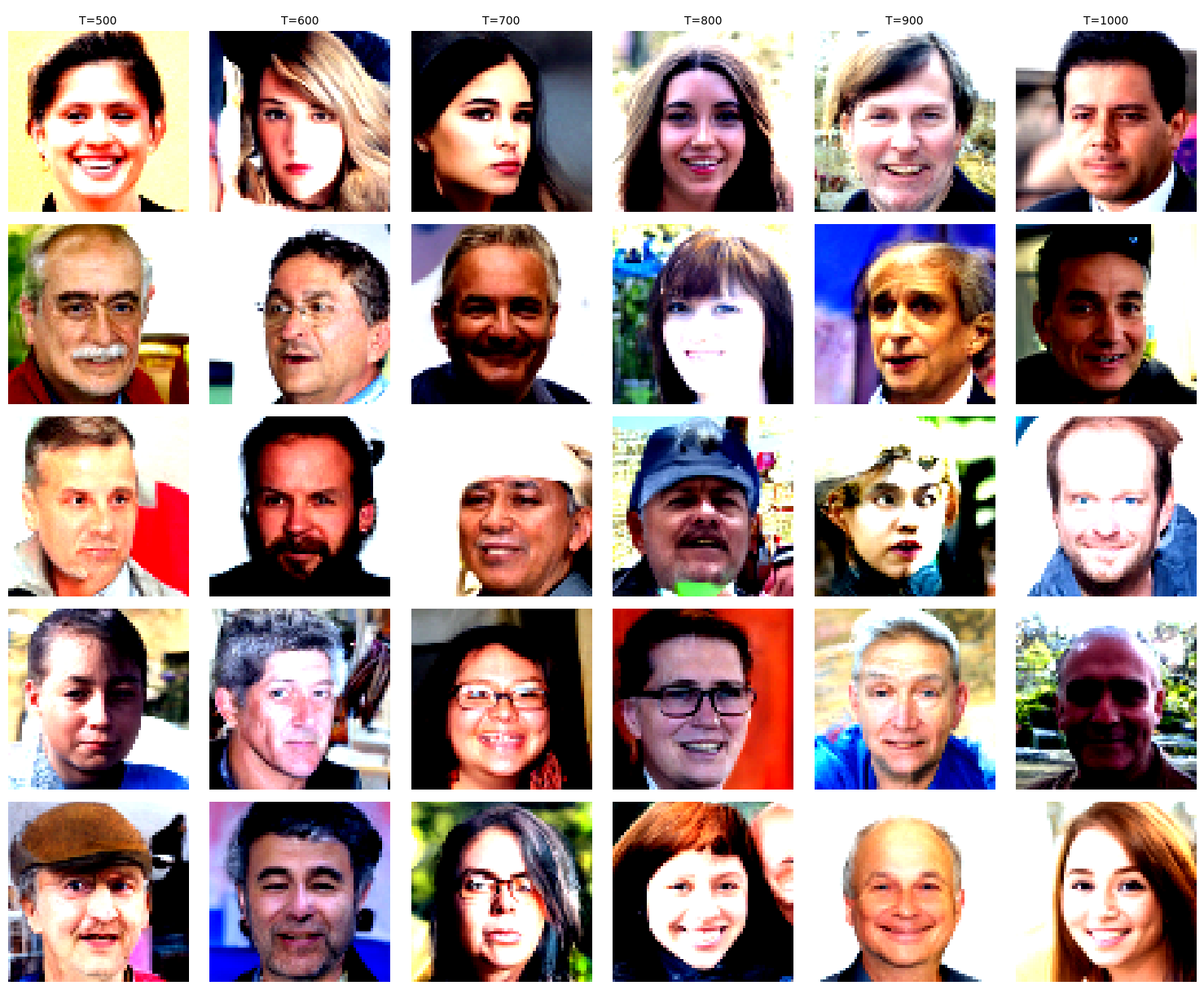}
    \caption{ Examples of the images generated with CDPM. Each column corresponds to one $T \in \{500, 600, 700, 800, 900, 100\}$.}
    \label{fig:cdpm_generated_images}
\end{figure*}
\\
\noindent\textbf{Effect of the number of integration steps.}
We vary here the number of sampling steps $\nsteps$ for models trained on Biased MNIST and BFFHQ and for Stable Diffusion.
In Fig.~6b and \cref{fig:2clbmnist_0.995_varnsteps_smin0.05_smax50_schurn40} 
we observe that in the 10-classes Biased MNIST $\rhomodel$ increases as $\nsteps$ increases (and likewise in \cref{fig:2clbmnist_0.995_varnsteps_smin0.05_smax50_schurn40}). More specifically, we can observe a \textit{reduction} of bias at low $\nsteps$, followed by an \textit{amplification} of bias at high $\nsteps$. The same is also evident in BFFHQ (\cref{fig:BFFHQ_rho0.95_varNsteps_smin0.01_smax80_schurn80}) and in Stable Diffusion (Fig.~6a). Furthermore, in \cref{fig:2clbmnist_0.995_varnsteps_smin0.05_smax50_schurn40} we remark that the range of values that $\rhomodel$ can take when we only vary $\nsteps$ is considerable: from $0.46$ to $0.74$. Therefore, the output distribution changes significantly, at least with respect to the bias, even though the quality of the generated digits does not change much as we see in \cref{fig:10cl_rho0.7_samples_varnsteps_samexinit}. 
\cref{fig:BFFHQ_rho0.95_varNsteps_smin0.01_smax80_schurn80} shows the same trend on BFFHQ as previously observed on Biased MNIST: at low $\nsteps$ there is reduction of bias and at high $\nsteps$ the bias is amplified.
Similarly, Fig.~6a shows the same increasing trend of $\rhomodel$ in Stable Diffusion. The trend remains when we use other values of $\eta$ (see \cref{fig:SD_portraitlawyer_eta0_0.3_0.7_varnsteps}). 
Since we do not have $\rhodataset$ as a baseline, we cannot conclude whether there is bias amplification, but we can at least observe that $\rhomodel$ varies significantly with $\nsteps$ (from $0.949$ to $0.987$), which is a previously unexplored phenomenon. In Fig.~6a we observe that although the HPSv2 score initially increases with $\nsteps$, it plateaus quickly, while $\rhomodel$ continues to increase.
\\\noindent\textbf{Time window with fresh noise.}
The two time windows in which the injected noise does not significantly change the generated distribution are $[40, 80]$ and $[0,5]$:
\begin{itemize}
    \item at the highest noise levels comprising $[40, 80]$, the bias (background color) may not yet be decided, hence the eventual variations due to stochasticity do not impact the generated color, 
    \item at the lowest noise levels comprising $[0,5]$, the characteristic features of the image (its digit and its color) have already appeared, and it remains only to refine the details.
\end{itemize}
Looking at the history of the denoised images in \cref{fig:10clBMINST_rho0.9_samples_history_noise}, we see that it is indeed possible to guess the end color starting from the time step $t\approx15$. However, this might be an effect specific to Biased MNIST, as we did not replicate the experiment with other datasets and models.\\
\noindent\textbf{The bias appears early.} Overall, our interpretation of how the generation process produces biased images is that the bias already appears in early stages. As such, introducing noise into the sampling process during these stages can help reduce the upsurge of bias in generated images. We present, in Fig.~\ref{fig:10clBMINST_rho0.9_samples_history_noise}, a qualitative visualization of how the denoising process produces samples in Biased MNIST (since the bias is the background color, it is easy to visually inspect). Indeed, the background color is the first generated feature, which supports our hypothesis. This could inspire more research in the field, notably conditioning the generation of non-biased attributes in the early stages.
\\\noindent\textbf{Robustness of the bias oracle.}
Since we use an oracle to obtain the gender and age labels of generated images on BFFHQ, we verify that this oracle is reliable by evaluating its accuracy on the train set. We also estimate the robustness of our oracle by classifying the noisy versions of the training data. To inject varying levels of noise, we add Gaussian noise with five different variances in the latent space of VQ-VAE in Stable Diffusion. The results for the full train set are shown in Fig.~\ref{fig:oracle_cdpm_accuracy_total} and the group-wise accuracies are shown in Fig.~\ref{fig:oracle_cdpm_accuracy_groups}, where each group corresponds to a pair of (gender, age) labels. We can see that the gender prediction remains robust to an arbitrary level of noise added in the latent space. As for age prediction, the overall performance remains relatively stable; however, we do observe a temporary drop in accuracy by $\sim 3\%$ for the medium level of noise. We speculate that the local features in this case might be affected too strongly, just enough to resemble the fine lines on the face, but not enough to affect the background. To support our assumptions, we provide the noisy images for all levels of noise considered in Fig.~\ref{fig:noisy_train_samples}. As a result, the oracle makes more mistakes in the younger groups. If we continue to add the noise, the background also becomes perturbed because the injected noise is too large, and the local features are better preserved. We provide examples of the images generated with CDPM in Fig.~\ref{fig:cdpm_generated_images} for reference.

{\small
\bibliographystyle{ieee_fullname}
\bibliography{egbib}
}



\end{document}